\documentclass[11pt]{article}   	
\usepackage[margin=.8in]{geometry}                		
\usepackage{graphicx}				
\usepackage{amssymb}
\usepackage{color}
\usepackage{amsmath}
\usepackage{appendix}
\usepackage{enumerate}
\usepackage{bbm}
\usepackage{mathabx}
\usepackage{hyperref}
\usepackage{amsthm}
\usepackage{authblk}
\usepackage[numbers]{natbib}

\usepackage{tikz}
\usetikzlibrary{arrows.meta}

\newtheorem{definition}{Definition}
\newtheorem{property}{Property}
\newtheorem{conjecture}{Conjecture}

\title{Capacity allocation analysis of neural networks: A tool for principled architecture design}
\author{Jonathan Donier\footnote{jdonier@spotify.com}}
\affil{Spotify}

\begin{document}
\maketitle

\begin{abstract}
Designing neural network architectures is a task that lies somewhere between science and art. For a given task, some architectures are eventually preferred over others, based on a mix of intuition, experience, experimentation and luck. For many tasks, the final word is attributed to the loss function, while for some others a further perceptual evaluation is necessary to assess and compare performance across models. In this paper, we introduce the concept of \emph{capacity allocation analysis}, with the aim of shedding some light on what network architectures focus their modelling capacity on, when used on a given task. We focus more particularly on \emph{spatial capacity allocation}, which analyzes \textit{a posteriori} the effective number of parameters that a given model has allocated for modelling dependencies on a given point or region in the input space, in linear settings. We use this framework to perform a quantitative comparison between some classical architectures on various synthetic tasks. Finally, we consider how capacity allocation might translate in non-linear settings.

\end{abstract}

\section{Introduction}

Since the popularization of deep neural networks in the early 2010s, tailoring neural network architectures to specific tasks has been one of the main sources of activity for both academics and practitioners. 
Accordingly, a palette of empirical methods has been developed for automating the choice of neural networks hyperparameters (a process sometimes called Neural Architecture Search), including -- but not limited to -- random search \cite{bergstra2011algorithms, bergstra2012random}, genetic algorithms \cite{miller1989designing, kitano1990designing}, bayesian methods \cite{springenberg2016bayesian, kandasamy2018neural} or reinforcement learning \cite{zoph2016neural}. However, when the computational requirements for training a single model are high, 
such approaches might be too expensive or result in iteration cycles that are too long to be practically useful -- though some work in that direction has been carried out recently \cite{domhan2015speeding, klein2016fast}. In other cases, when the loss function is only used as a proxy for the task at hand \cite{van2016wavenet,  van2016conditional, jansson2017singing} or is not interpretable \cite{goodfellow2014generative}, a further perceptual evaluation is typically necessary to evaluate the quality of a model's outputs and such systematic approaches at least partially break down. In both cases, an efficient and quantitative method to analyze and compare neural network architectures would be highly desirable -- be it only to come up with a limited set of plausible candidates to pass on to the more expensive (or manual) methods.


In this paper, we introduce the notion of \emph{capacity allocation analysis}, which is a systematic, quantitative and computationally efficient way to analyze neural network architectures by quantifying which dependencies between inputs and outputs a parameter of a set of parameters actually model. We develop a quantitative framework for assessing and comparing different architectures for a given task, providing insights that are complementary to the value of the loss function itself. 

In this paper, we develop the theory in linear settings, where both models and data are linearized. Linearizing neural networks might be regarded as the inverse of ``neuralizing'' linear models: while the intuition for the latter is to augment the desirable properties of some well-know (often linear) method with the expressivity \cite{raghu2016expressive, poole2016exponential, guss2018characterizing} of deep and non-linear neural networks, linearizing neural networks provides a way to quantify some of the properties that might be characteristic of a given architecture through theoretical analysis. In some way, both approaches make the leap of faith that some properties remain more or less valid independently from the complexity of the data and the expressivity of the model (in particular, its degree of non-linearity). 

We focus more particularly on \emph{spatial} capacity allocation, with the following intuition: a network's spatial architecture (i.e. whether it uses fully connected, recurrent, convolutional, dilated layers, etc.) tends to define its capacity allocation across the input space, while its complexity (its non-linearities, number of channels, etc.) tends to define the complexity of the dependencies that it can model. As hinted above, we mostly set aside the latter for now and choose to focus on the spatial aspect of neural network architectures by considering linearized versions of the models we wish to analyze. How much the two dimensions of the problem can be disentangled remains to be understood, 
 but this is the leap of faith that we are willing to make here.

Our work is related to \textit{a posteriori} analysis of trained models (also sometimes called \emph{network inspection}), which has been the object of many recent studies \cite{simonyan2013deep, zeiler2014visualizing, zintgraf2017visualizing, sellam2018deepbase}, as a way to peek into the neural black boxes. The goal of most of this literature is to analyze a network's \emph{activations}, to understand which property of the input lead to or correlate with a given behaviour -- for example, a final classification decision or simply the activation of a particular unit in the network. Such methods however differ from the present one in that they are mostly \emph{example-based} rather than intrinsic, i.e. they mostly make sense on particular instances of the input rather than in general. In contrast, we are looking for an objective, quantitative and computationally efficient way to compare model architectures (rather than specific \emph{instances} of such architectures that are obtained after expensive training) to provide grounds for principled network design.

We start by providing some guiding intuitions in Section \ref{sec:intuitions} before introducing formally the concept of capacity allocation in linear systems in Section \ref{sec:capacity}. We introduce the notion of spatial capacity allocation by showing that capacity can be broken down along subspaces of the input space, and show that capacity can be used to provide statistical upper bounds on the model error. We also introduce the notion of conditional capacity, which allows to study the conditional influence of subsets of constraints. Section \ref{sec:applying_capacity} then applies capacity allocation analysis in the case of linear(ized) models and linear tasks (e.g. gaussian process prediction tasks). We show that the \emph{total model capacity} $\kappa$ at some optimal state corresponds to its effective number of parameters, and that it can be broken down across its input space and its parameters (or layers). 
Section \ref{sec:examples} illustrates the theory on two common type of architectures -- hierarchical and recurrent -- and presents several insights in both cases. Finally, Section \ref{sec:multidim} considers how capacity allocation might translate in non-linear settings.

\section{Guiding intuitions}\label{sec:intuitions}


Our end goal is to define some \emph{extensive property}\footnote{In physics, an extensive property is a property which is additive for subsystems, while an intensive property is a property that does not depend on the system size.} which characterizes a model's modelling capacity, and to use it to perform comparisons between models in a meaningful and objective way for a given task. We wish to be able to break down this quantity (which we call the \emph{capacity}, denoted $\kappa$) across various dimensions or subspaces of the input space -- for example, the spatial dimensions -- to quantify how much of this capacity is allocated for each subspace: this is the \emph{spatial capacity allocation} alluded to above. 

\begin{figure}[!t]
\includegraphics[width=\textwidth]{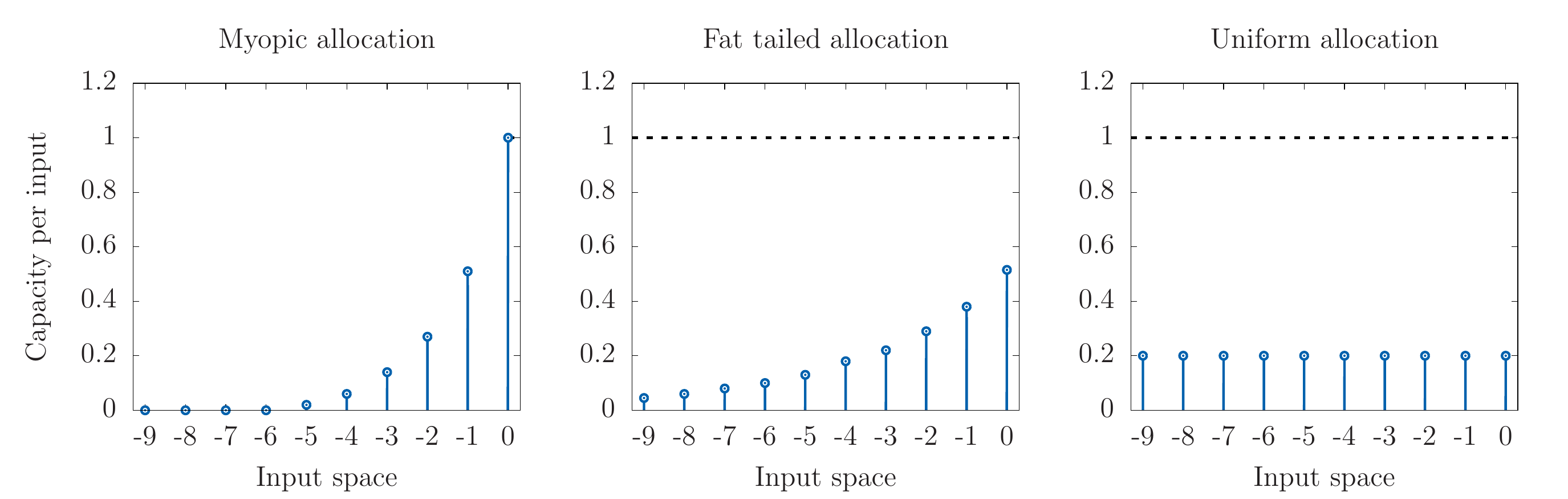}
\centering
\caption{\label{fig:schema} A synthetic example of spatial capacity allocation for 3 models on the same task, which involves a prediction from a 1-dimensional input process. The $x$ axis represents the position in the input space, while the $y$ axis represents the spatial capacity allocation. The area under the curve is the same for all 3 models, but their spatial allocation differs.}
\end{figure}

The appeal of such a quantity is particularly salient when the loss function is only a proxy for the task one want to achieve -- which is often the case with generative models. In that case, minimizing the loss might result in spurious behaviour which is suboptimal from the point of view of the task. 
Such considerations must therefore be taken into consideration earlier in the design process. One example, which has been one motivation for the present work, is the task of artificial music generation using autoregressive models \cite{van2016wavenet}, which are usually trained in a 1-step-ahead fashion, hoping that long term dependencies will accessorily be captured to produce a musical output -- rather than mere babbling. Such tradeoff between audio quality and structure has been described in \cite{dieleman2018challenge}, whose authors point out that lower audio quality might be the price to pay to be able to capture long term structure -- so they voluntarily limit the former to gain on the latter. One goal of the present theory is to make sense of such observation, and guide network design in a principled way: in the terms of capacity allocation analysis, one would want in their case to allocate enough capacity to remote inputs rather than focussing on the recent past, even though this might be suboptimal for the loss considered. At this stage, this is just a construction of the mind, but our goal is to make this intuition quantitative. 

Figure \ref{fig:schema}  shows some fictitious spatial capacity curves for three different models, whose task is to make some prediction from $n=10$ inputs represented on the $x$ axis (this can be thought of as predicting the next sample of a 1-dimensional autoregressive process). Each model has a fixed intrinsic capacity (the area under the curve), which is being allocated spatially in a way that might be specific to the task at hand (i.e. the joint distribution of the input and output variables). For the same task, different architectures might focus on different dependencies -- in the example, Model 1 focuses on the far right part of the space (i.e. the most recent past, in the autoregressive case), while Model 2 looks further onto the left and Model 3 looks uniformly at the whole input space. In the context of the music generation example mentioned above, one might thus prefer Models 2 or 3 over Model 1, as they are more susceptible to capture long term structure. As we will see, this pattern of excessive capacity allocation to the recent past is in fact typical of time series prediction, and one major challenge associated with multi-scale tasks such as audio modelling.

More generally, capacity analysis might be used (i) to tailor architectures to specific needs, as in the case just mentioned, or (ii) to simply analyze and compare architectures \textit{a posteriori} to get a better understanding of what they achieve (see Appendix \ref{sec:wavenet_repeat} for an example in the context of Wavenets).



\section{Capacity allocated to a subspace} \label{sec:capacity}

Before defining capacity allocation in the context of linear models (which will be the topic of Section \ref{sec:applying_capacity}), we start by defining the concept more generally for linear systems.

\subsection{Total capacity}

Let us consider a linear system, i.e. a set of $\kappa$ linear orthonormal constraints: 

\begin{equation}\label{eq:linear_system}
K^TX = Y ,\quad 
  \begin{cases} 
   X \in \mathbb{R}^n &\text{(variable)} \\
   K \in \mathbb{R}^{n\times\kappa}& \text{(constraints)}\\
   Y \in \mathbb{R}^\kappa &\text{(target)}
  \end{cases}
\end{equation}

The orthonormality requirement means that the columns of $K$ (i.e. the constraints) are orthonormal vectors in $\mathbb{R}^n$. In this linear setting, we will call the number of (orthogonal) constraints, the \emph{total capacity} $\kappa$. Naturally, each additional constraint decreases the dimension of the space of possible values that $X$ can take. Note that in a space of dimension $n$, the subspace that satisfies $\kappa$ independent linear constraints has a dimension $\delta = n - \kappa$, often called the number of \emph{degrees of freedom} in statistics\footnote{We prefer to use the concept of capacity rather than degrees of freedom, as this concept will be better suited to analyze neural networks later on.} \cite{box1978statistics}. At full capacity $\kappa=n$ (equivalently, $\delta=0$ degrees of freedom), $X$ is fully constrained and is equal to $\left(K^T\right)^{-1}Y$. 


\subsection{Spatial capacity}

We now want to be more specific and define the notion of \emph{capacity allocated to a subspace}, to quantify how many constraints are being applied along a given subspace of the input space.


\begin{definition}\label{def:capacity}
Let $\mathcal{K}$ be a vector subspace of $\mathbb{R}^n$ of dimension $\kappa$. Let $K = \{k_1, ..., k_\kappa\} \in \mathbb{R}^{n\times\kappa}$ be an orthonormal basis of $\mathcal{K}$, and $\mathcal{K}^\perp$ the orthogonal complement of $\mathcal{K}$, defined by the set of points that satisfy the linear constraints $\{X \in \mathbb{R}^n \mid K^TX = Y \}$. Let  $\mathcal{S}$ be a vector subspace of $\mathbb{R}^n$ with orthonormal basis $S = \{s_1, ..., s_{n_S}\}  \in \mathbb{R}^{n\times n_S} $. We define the \emph{capacity allocated to subspace $\mathcal{S}$ by $\mathcal{K}$}, noted $\kappa_\mathcal{K}(\mathcal{S})$ as the Frobenius norm of the matrix $K^T S$:
\begin{equation}
\kappa_\mathcal{K}(\mathcal{S}) := \| K^T S  \|_F^2 =   \mathrm{Tr} \left( S^T K K^T S \right).
\end{equation}
\end{definition}

For simplicity, when there is no ambiguity regarding the set of constraints, we will omit the subscript and use the notation $\kappa(\mathcal{S})$. The capacity allocated to a subspace has a number of convenient and intuitive properties, which we detail below:

\begin{property}
Since the Frobenius norm is rotation- and permutation-invariant, the capacity allocated to a subspace $\mathcal{S}$ by $\mathcal{K}$ does not depend on the particular orthonormal bases that are chosen to represent $\mathcal{S}$ and $\mathcal{K}$. The capacity is therefore an \emph{intrinsic} property of $\mathcal{S}$ and  $\mathcal{K}$.
\end{property}

\begin{property}
If $\mathcal{S}  = \mathbb{R}^n$, then the orthonormality of $K$ gives $\kappa(\mathcal{S}) = \| K \|_F^2 = \kappa$: the capacity allocated to the whole space is equal to the number of independent constraints $\kappa = \mathrm{dim}(\mathrm{Vect}(K))$.
\end{property}

\begin{property}
The capacity allocated to the vector subspace generated by two orthogonal vector subspaces $\mathcal{S} = \mathcal{S} _1 \obot \mathcal{S} _2$ is equal to the sum of their allocated capacities: $\kappa(\mathcal{S})  = \kappa(\mathcal{S}_1)  + \kappa(\mathcal{S}_2) $. In particular, if $\{s_1, ..., s_n\}$ is an orthonormal basis of $\mathbb{R}^n$, then the sum of the capacities allocated to each subspace $\mathrm{Vect}(s_i)$ is equal to the capacity allocated to the whole space: 
\begin{equation}
\sum_{i=1}^n\kappa(\mathcal{S}_i)  = \kappa(\mathcal{S}) .
\end{equation}
\end{property}

\begin{property}
The capacity allocated to the 1-dimensional subspace $\mathrm{Vect}(k)$, where $k\in \mathrm{Vect}(K)$ belongs to the set of constraints, is equal to 1. The capacity allocated to the subspace $\mathrm{Vect}(k^\perp)$, where $k^\perp\in \mathrm{Vect}(K)^\perp$ is orthogonal to the set of constraints, is equal to 0.
\end{property}

Basically, the capacity allocated to a given subspace $\mathcal{S}$ represents the number of independent constraints that are being used to constrain the projection of $X\in \mathcal{K}^\perp$ onto that subspace. For every vector $k\in\mathcal{K}$ in the space of constraints, the above property means that exactly one constraint is being used to enforce the constraint $k.X = y$. For every vector $k^\perp\in\mathcal{K}^\perp$ orthogonal to the set of constraints, the projection $k^\perp .X$ is unconstrained (it uses 0 constraints). 

For a given orthonormal basis $\{s_1, ..., s_n\}$ of $\mathbb{R}^n$, the respective capacities $\{\kappa(s_1) , ..., \kappa(s_n) \}$ represent the respective numbers of independent constraints (between 0 and 1) that are being used to constrain $X$ along each axis. These capacities sum to the number of free parameters $K$ in the model. In short, the notion of capacity allocated to subspaces allow us to break down how the constraints are being allocated with respect to a given partition of the space.

To give some flesh to the above ideas, let us jump ahead and anticipate Section \ref{sec:applying_capacity}, where $X$ will typically represent the difference between a linear model's effective AR coefficients and the true AR coefficients of some gaussian process. Each of the model parameters will give rise to exactly one (linear) constraint through the chosen optimality criterion. If the model allocates 1 free parameter to enforce a constraint along a given dimension, then the modelling error along that dimension will be zero. Analyzing the capacity allocated per dimension will thus allow us to understand which components of the AR coefficients are being captured, and to which extent. By definition, a linear model with a number of free parameters that matches the dimension of the input space (capacity $\kappa=n$, or equivalently number of degrees of freedom $\delta=0$) will be able to reproduce exactly all the true coefficients as the error will be fully constrained, while an underparametrized model ($\kappa<n$, $\delta>0$) will have to allocate its sparse resources to a larger number of coefficients -- and choose which ones to put more capacity on.

\subsection{Statistical bounds on errors}\label{sec:errors_distribution}

Given some capacity allocation along a given subspace, can we derive bounds on the errors\footnote{We are again jumping ahead and assuming by using this terminology that $X$ will correspond to errors with respect to the true model such that the ideal state is $X=0$, cf Section \ref{sec:applying_capacity}.} $X$ along that same subspace? Let us consider the constraint $K^TX = 0$ with $K = \{k_1, ..., k_\kappa\} \in \mathbb{R}^{n\times\kappa}$, and $K^\perp = \{k_{\kappa + 1}, ..., k_n\} \in \mathbb{R}^{n\times\left(n - \kappa\right)}$ a complementary basis of $\mathbb{R}^n$ such that $\mathrm{Vect}(K) \bigoplus \mathrm{Vect}(K^\perp) = \mathbb{R}^n$. A given vector of errors $X$ can be written as:

\begin{equation}\label{eq:error_expression}
X = \sum_{i=\kappa + 1}^n \lambda_i k_i,\quad \forall i,  \lambda_i\in\mathbb{R}
\end{equation}

\noindent Let us define the error $e_S$ along some subspace $\mathcal{S}=\mathrm{Vect}(S)$ of dimension $n_S$ as $e_S = \| X^T S  \|_F$, where $S$ is again an orthonormal basis, as well as the average squared error:

\begin{equation}\label{eq:error}
\epsilon_S := \mathbb{E}_\lambda \left(e_S^2\right) =\mathbb{E}_\lambda\left( \| X^T S  \|_F^2\right),
\end{equation}

\noindent where the expectation is taken over the distribution of $\lambda$'s.\footnote{We will see later than the values of $\lambda$ are constrained by the model space. Thus averaging over $\lambda$  can be seen as ``averaging over model spaces''.} This expectation is not tractable in general,\footnote{...and meaningless in general without further assumptions.} but by assuming some symmetry in $\lambda$ (such that $\lambda_i$'s are i.i.d. of variance $\sigma_\lambda^2$), some elementary manipulations then give:

\begin{equation}\label{eq:error_bound}
\begin{aligned}
\epsilon_S &=\mathbb{E}_\lambda\left( \| X^T S  \|_F^2\right) = \mathbb{E}_\lambda\left( \| \sum_{i=\kappa + 1}^n \lambda_i k_i^T S  \|_F^2\right) = \sigma_\lambda^2  \sum_{i=\kappa + 1}^n \| k_i^T S  \|_F^2\\ 
&= \sigma_\lambda^2 \| \left(K^\perp\right)^T S  \|_F^2\\
&= \sigma_\lambda^2 \left(n_S - \kappa_S \right) =\frac{n_S - \kappa_S}{n - \kappa}  \epsilon_{\mathbb{R}^n}\\
&= O\left(n_S - \kappa_S \right)
\end{aligned}
\end{equation}



\noindent This means that the squared error along a given subspace $\mathcal{S}$ of dimension $n_S$ is statistically bounded by the dimension of that subspace minus the capacity allocated by the model to that subspace. If the model allocates a full capacity to a subspace, $\kappa_S = n_S$ and therefore $\epsilon_S = 0$: that subspace is perfectly modelled. 

This calculation only gives a statistical order of magnitude of the errors, for a fixed capacity allocation. In real settings, model spaces often have specific structures rather than ``average'' ones, such that the errors can differ greatly from the statistical bound. In particular, some true coefficients are often vanishingly small in practice, which leads to vanishingly small errors in spite of a zero or near-zero capacity allocation. Eq. (\ref{eq:error_bound}) should therefore be understood as a statistical upper bound, rather than viewed as a general equality. In fact, this is precisely what makes the capacity theory appealing, as it quantifies the modelling capacity allocated to modelling some given dependencies (or conversely the degree of freedom allowed), regardless of the realized (and idiosyncratic) errors, which are input- and model-space-dependent.

There is however one case where the equality holds exactly, i.e. where it is not necessary to take the expectation in Eq. \ref{eq:error} to get closed form results. Indeed when $\kappa=n - 1$, there is only one element in the sum in Eqs. (\ref{eq:error_expression}) and (\ref{eq:error_bound}) and one can write directly:

\begin{equation}
e_S^2 = \lambda_n^2 (n_S - \kappa_S) = (n_S - \kappa_S) e_{\mathbb{R}^n}^2, \quad \quad \mathrm{where}~  e_{\mathbb{R}^n}^2 = \| E  \|_2^2, \quad 0\leq n_S - \kappa_S \leq 1.
\end{equation}

\noindent In that case, the relative squared errors along subspaces $e_S^2 / e_{\mathbb{R}^n}^2$ are \emph{exactly} the complementary of the corresponding capacities.

\subsection{Conditional capacity} \label{sec:conditional_capacity}

The concept of capacity is analogous to that of probabilities, in that this is an object that can only be defined jointly for a set of constraints. In particular, the sum of the capacities allocated by two sets of constraints is not equal in general to the capacity allocated by the direct sum of these two constraint spaces, unless these constraint spaces are orthogonal (this is akin to independence in probabilities):

\begin{property}\label{prop:sum}
Let $\mathcal{K}_1$ and $\mathcal{K}_2$ be two non-trivial spaces of constraints, and $\mathcal{K}=\mathcal{K}_1\bigplus \mathcal{K}_2$. Then the following equivalence holds:

\begin{equation}
\left( \forall \mathcal{S},~ \kappa_{\mathcal{K}_1}(\mathcal{S}) + \kappa_{\mathcal{K}_2}(\mathcal{S}) = \kappa_{\mathcal{K}}(\mathcal{S}) \right) \quad \Leftrightarrow \quad \mathcal{K}=\mathcal{K}_1 \obot \mathcal{K}_2.
\end{equation}

\end{property}

\noindent The proof of the above equivalence is presented in Appendix \ref{sec:proof}. We therefore define the \emph{conditional capacity} allocated by a set of constraints $\mathcal{K}_1$ given a set of constraints $\mathcal{K}_2$:

\begin{definition}
Let $\mathcal{K}_1$ and $\mathcal{K}_2$ be two spaces of constraints. The \emph{conditional capacity} allocated by $\mathcal{K}_1$ to a vector subspace $\mathcal{S}$ \emph{given} another space of constraints $\mathcal{K}_2$, noted $\kappa_{\mathcal{K}_1}\left(\mathcal{S} \mid \mathcal{K}_2 \right)$, is defined as:

\begin{equation}
\kappa_{\mathcal{K}_1}\left(\mathcal{S} \mid \mathcal{K}_2 \right) := \kappa_{\mathcal{K}_1 \bigplus \mathcal{K}_2}(\mathcal{S}) - \kappa_{\mathcal{K}_2}(\mathcal{S})
\end{equation}

\end{definition}

\noindent This quantifies the additional capacity that $\mathcal{K}_1$ brings over $\mathcal{K}_2$. If $\mathcal{K}_1 \subset \mathcal{K}_2$ for example, the conditional capacity is equal to zero as no new constraints are being added. As in probabilities, this definition gives rise to various properties, among which the following identity on chains of conditional capacities:

\begin{property}\label{prop:conditional}
Let $\left\{\mathcal{K}_i\right\}_{i=1,\ldots,n}$ be $n$ spaces of constraints and $\mathcal{K} = \bigplus_{i=1}^{n} \mathcal{K}_i$. Then the following holds:

\begin{equation}
\forall \mathcal{S},~ \kappa_{\mathcal{K}}(\mathcal{S}) = \sum_{i=1}^n \kappa_{\mathcal{K}_i}(\mathcal{S} \mid \bigplus_{j=1}^{i-1} \mathcal{K}_j  )
\end{equation}

\end{property}

\noindent Note that this is akin to the chain rule of probabilities. This will allow later on to decompose a model's capacity into the (conditional) capacities of each of its parameters, or each of its layers (see e.g. Section \ref{sec:wavenet_further}).

\section{Capacity applied to linear models}\label{sec:applying_capacity}

In the previous section, we have defined the concept of capacity in the abstract. Here, we show how it applies to the case of trained linear models, by allowing one to determine \textit{a posteriori} what dependencies a given model has captured once it has reached a (locally) optimal state. More precisely, we want to determine what part of the input space the model has focused its modelling capacity on, by determining which components are tightly fixed and which ones are free to vary -- in a quantitative manner. Since one of the main tasks of model architecture design is to impose which dependencies between its inputs and outputs the model should try to capture, this framework should be useful for approaching the task in a more principled way.

As we will see, one can map a model's parameters with a corresponding set of linear constraints, such that the capacity of the \emph{model}, defined as the capacity $\kappa$ associated with its associated set of constraints, is equal to its number of free parameters. For a given subspace $\mathcal{S}$ of the input space, the model's capacity allocated to $\mathcal{S}$ then quantifies \emph{how many free parameters} it allocates for reproducing dependencies along that subspace. In particular, this will allow us to define a model's \emph{spatial capacity allocation}, as its capacity allocation along the natural dimensions of the input space.

\subsection{Models manifold}

Let us start by defining some terminology related to linear models, which will be useful throughout the rest of the paper. We consider linear models with 1-dimensional outputs:

\begin{equation}
\begin{aligned}
A: \quad& \mathbb{R}^n & \to &  \mathbb{R}\\
 & X &  \mapsto & A^TX.
\end{aligned}
\end{equation}

\noindent The \emph{model space} (i.e. the ensemble of possible values of $A$) is defined by some parametrization:

\begin{equation}
\begin{aligned}
 f: \quad& \mathbb{R}^p & \to & \mathbb{R}^n\\
    & W & \mapsto & A_W
    \end{aligned}
\end{equation}

\noindent which defines the space of models as $\mathcal{A} = f( \mathbb{R}^p)\subset \mathbb{R}^n$ from a parameter space $\mathcal{W} = \mathbb{R}^p$. 

The components of $W\in\mathcal{W}$ are called the model \emph{parameters}, while the components of $A\in\mathcal{A}$ are called the model \emph{coefficients}. Typically, the space of models $\mathcal{A}$ will be a $\kappa$-dimensional manifold where $\kappa \leq p$ represents the number of effective parameters of the model (\textit{aka} the total model capacity). We also define the space of \emph{errors} with respect to some target model $A^*$ as:

\begin{equation}
\mathcal{X} = \mathcal{A} - A^* := \{A - A^*  \mid A \in \mathcal{A} \},
\end{equation}

\noindent and accordingly we will note the model error $X := A - A^*$.

\subsection{Optimization program}\label{sec:subspaces}

Assume that one tries to learn some target model $A^*$ by minimizing some quadratic loss over some model space $\mathcal{A}$:

\begin{equation}
\begin{aligned}\label{eq:optim_A}
& \underset{A\in \mathcal{A}}{\text{argmin}} ~ &\mathcal{L} := \left(A - A^*\right) ^T \Sigma \left(A - A^*\right) \\
 \Leftrightarrow \quad & \underset{X\in \mathcal{X}}{\text{argmin}} ~& X ^T \Sigma X,
\end{aligned}
\end{equation}

\noindent or in the parameter space,

\begin{equation}
\begin{aligned}\label{eq:optim_W}
 \underset{W\in \mathbb{R}^p}{\text{argmin}} ~ \mathcal{L} := \left(A_W - A^*\right) ^T \Sigma \left(A_W - A^*\right).
\end{aligned}
\end{equation}

\noindent For example, one might be trying to predict the next sample of some gaussian process with lag-$n$ covariance matrix $\Sigma$ using a linear model with a receptive field of size $n$, parametrized by $W$ (the expression for the optimal model $A^*$ in that case is provided in Appendix \ref{sec:optimization}). In the case where the model space $\mathcal{A}$ is a $\kappa$-dimensional manifold (i.e. parametrized by $\kappa$ independent parameters), selecting one model in $\mathcal{A}$ (or equivalently, one error in $\mathcal{X}$) requires to impose $\kappa$ independent constraints on the system, which will stem from the optimality criterion. An optimal model  therefore:

\begin{enumerate}[(i)]
\item belongs to the space of models $\mathcal{A}$ parametrized by $W$,
\item satisfies a set of orthonormal linear constraints $K\in{\mathbb{R}^{n\times\kappa}}$ imposed by the optimality criterion (which are task-dependent).
\end{enumerate}

The first condition is imposed by the parametrization of the model space (for example some linearized neural network architecture), while the second condition describes the tradeoffs that the model has to make when modelling the input space -- i.e. which dependencies to focus on when allocating its parameters.

\begin{figure}[ht]
\includegraphics[width=0.55\textwidth]{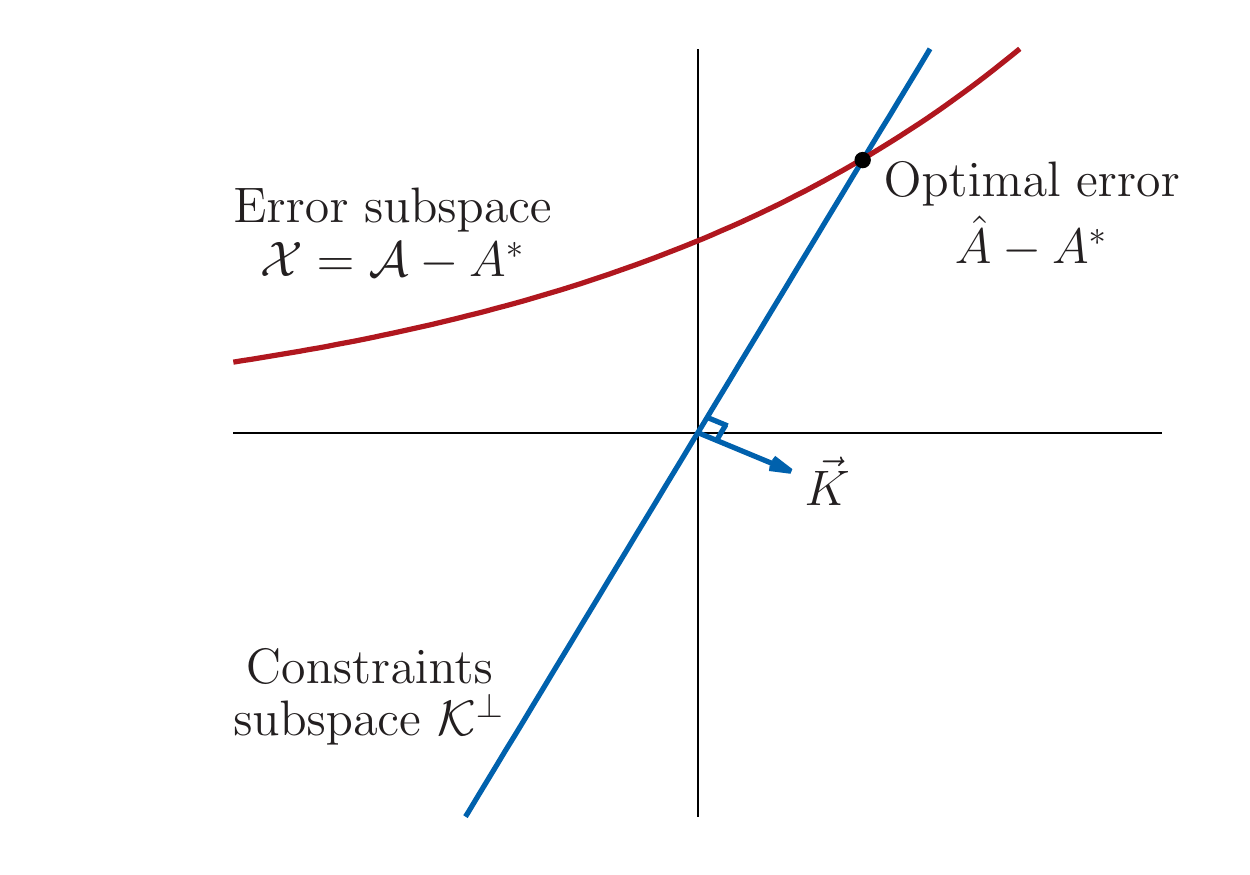}
\centering
\caption{\label{fig:subspaces} Example of an error subspace $\mathcal{X}$ and a constraints subspace $\mathcal{K}^\perp$ for a 2-dimensional input space and a 1-dimensional model manifold (i.e. $\mathcal{X}$ is parametrized by a single parameter $w\in\mathbb{R}$). The intersection between the two manifolds is reduced to one point here, which is the only result of the optimization program in Eq. \ref{eq:optim_A}.}
\end{figure}

The intersection of the errors manifold $\mathcal{X} := \mathcal{A} - A^*$ (of dimension $\kappa$) with the orthogonal of the constraints subspace, $\mathcal{K}^\perp$ (of dimension $n - \kappa$), will then give us a set of locally optimal errors, and therefore a set of locally optimal models. If the optimization program has only one local minimum equal to the global minimum, the intersection will be reduced to the singleton containing the optimal error: $\mathcal{X}\cap \mathcal{K}^\perp = \{\hat{A} - A^* \}$.  A graphical representation is shown in Fig. \ref{fig:subspaces} for a 2-dimensional input space and a 1-dimensional model manifold. 

Our goal below will be to determine the set of $\kappa$ orthonormal linear constraints $K \in \mathbb{R}^{n\times\kappa}$ derived from the above optimization program. We will then be able to perform a capacity analysis of the (locally) optimal model using the tools introduced in Section \ref{sec:capacity} by considering the constraints space $\mathcal{K}$.

\subsection{Constraints subspace at a locally optimum state}\label{sec:constraints}

Let $\hat{W} \in \mathbb{R}^{p}$ be a set of parameters that achieves a local optimum of $\mathcal{L}$ in Eq. (\ref{eq:optim_W}). Then the following relations hold at $W = \hat{W}$ :

\begin{equation}
\begin{aligned}\label{eq:optimum}
\frac{\partial \mathcal{L}}{\partial W} = 0 \quad & \Leftrightarrow \quad \left(\frac{\partial A}{\partial W}\right)^T \Sigma \left(A - A^* \right) = 0\\ 
& \Leftrightarrow \quad  \left(\Sigma \frac{\partial A}{\partial W}\right)^T . X= 0\\
& \Leftrightarrow \quad \tilde{K}^TX = 0, & \quad \text{where } \tilde{K} :=  \Sigma \frac{\partial A}{\partial W} \in \mathbb{R}^{n\times p}.
\end{aligned}
\end{equation}

\noindent One can find an orthonormal basis $K$ of $\mathrm{Vect}(\tilde{K})$ (the vector space generated by the columns of $\tilde{K}$) using the factorization of the Gram matrix $\tilde{K}\tilde{K}^T = Q\Lambda Q^T$, where $Q\in\mathbb{R}^{n\times n}$ is a rotation matrix and $\Lambda\in\mathbb{R}^{n\times n}$ is a positive diagonal matrix with $\kappa \leq n$ non-zero diagonal values, which we call the \emph{capacity weighting matrix} (note that its number of non-zero eigenvalues is equal to the number of effective parameters $\kappa$, which is a convenient method to compute $\kappa$). Then, define $K$ as the $n\times \kappa$ matrix containing the $\kappa$ columns of $Q$ that correspond to the non-zero eigenvalues. The above relations are then equivalent to:

\begin{equation}
K^TX = 0,
\end{equation}

\noindent where the columns of $K$ are orthonormal vectors and $\mathrm{Vect}(K)=\mathrm{Vect}(\tilde{K})$. These $\kappa$ constraints determine which coefficients of the (locally) optimal model are tightly imposed at the optimal point (number of degrees of freedom per dimension close to 0, or equivalently allocated capacity per dimension close to 1), and which ones are virtually free to vary (number of degrees of freedom per dimension close to 0, or equivalently allocated capacity per dimension close to 1).

For a given subspace $\mathcal{S}$ with orthonormal basis $S$, we can therefore compute the corresponding capacity allocated by the model according to Definition \ref{def:capacity}, as:

\begin{equation}
\kappa(\mathcal{S}) =  \| K^T S  \|_F^2 .
\end{equation}

\noindent One particularly interesting partition of the space we will consider below is the partition according to the natural basis of $\mathbb{R}^n$, which will allow us to perform a \emph{spatial capacity analysis} of our models, \textit{i.e.} to analyze their capacity allocation along the spatial dimensions of the input space, for a range of model architectures and input distributions. Another interesting study would be to perform a frequency analysis along Fourier components.

\section{Examples}\label{sec:examples}

We now illustrate the theory above on two types of architecture that are popular for modelling 1-dimensional data with long-range dependencies: hierarchical models and recurrent models. In both cases, we will consider the task of predicting the next sample of a gaussian process with autocorrelation process $C = \{c_i \}_{i \geq 0}
$ (equivalently, its autocorrelation matrix $\Sigma$ where $\Sigma_{ij} := c_{\mid i-j \mid}$), from its last $n$ inputs. The exact solution $A^*$ and the associated optimal variance $v^*$ for this problem are given in Appendix \ref{sec:GP}. 

\subsection{Hierarchical models}\label{sec:examples_hierarchical}

Hierarchical models have become popular since the introduction of Wavenets \cite{van2016wavenet, engel2017neural, oord2017parallel} for modelling audio signals, which are one prime example of 1D signals with long-range dependencies. Indeed, audio signals typically have tens of thousands of samples per second in order to cover the full spectrum of our auditory perception. In order to capture such long range dependencies while keeping a manageable number of parameters and reasonable memory requirements, the authors of \cite{van2016wavenet} have introduced Wavenets, which use a hierarchical architecture using dilated convolutions with exponentially growing dilation rate, resulting in a receptive field that grows exponentially in the number of layers. In this section, we investigate simplified linearized versions of such hierarchical models using the tools introduced in the previous sections, to see what properties of the input space they capture -- and what they focus their capacity on.

\subsubsection{Model definition}

\begin{figure}[!t]
\hspace*{-7.8cm}  \includegraphics[width=1.8\textwidth]{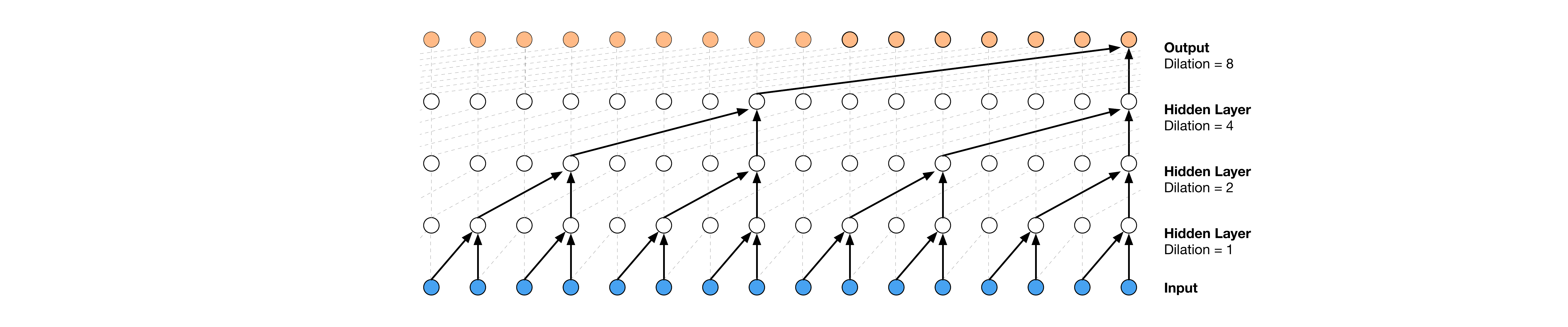}
\centering
\caption{\label{fig:wavenet} The architecture of a hierarchical model for $n_l = 2, d_l=2^{l-1}, c_l=1$ (image reproduced from \cite{van2016wavenet} with the permission of the authors).}
\end{figure}

The class of hierarchical linear models we consider here are the models of the form:

\begin{equation}
f(X) = W^{(L)} * W^{(L - 1)} * ... * W^{(1)} * X,
\end{equation}

\noindent where $*$ denotes the convolution operator. Each layer $1 \leq l \leq L$ consists of a filter of size $c_{l-1} * n_l * c_l$ and dilation rate $d_l$, where $c_l$ is the number of channels of layer $l$ and $n_l$ is the spatial extent of the filters at that layer. An example with $n_l = 2, d_l=2^{l-1}, c_l=1$ is represented in Fig. \ref{fig:wavenet}.

In this particular case where $d_l=n_l^{l-1}$ and $c_l=1$, the space of models $\mathcal{A}$ is parametrized as:

\begin{equation}\label{eq:space_hierarchical}
\mathcal{A} = \left\{ \left(a_i = \prod_{l=1}^L W^{(l)}_{\text{mod}(\left\lfloor \frac{n-i}{d_l}\right\rfloor, n_l * d_l)} \right) _{i=1..n} \mid  W^{(l)} \in \mathbb{R}^{n_l}, \forall l\in\left[1, L \right] \right\},
\end{equation}

\noindent where the total receptive field of the model is $n := n_L * d_L$. The above parametrization is such that every coefficient $a_i$ can be written as a product of $L$ coefficients, one from each layer. 
Note that in this case the mapping from parameters to models is affine in each of its inputs, since parameters are not shared across layers.

\subsubsection{Capacity analysis}

The space of models $\mathcal{A}$ defined by Eq. (\ref{eq:space_hierarchical}) is quite complex and finding closed form solutions is not an easy task, therefore we use numerical optimization over the parameters $W$ to find the optimal solution $\hat{A}$ to Eq. (\ref{eq:optim_W}). We can then perform a capacity analysis of the optimal model according to the theory developed in the previous sections, and compare the optimal variance $\hat{v}$ to the theoretical lower bound $v^*$ (note that they are related \textit{via} the loss function through the relation $\hat{v} = v^* + \mathcal{L}$). 

\begin{figure}[!t]
\hspace*{-1cm} \includegraphics[width=1.1\textwidth]{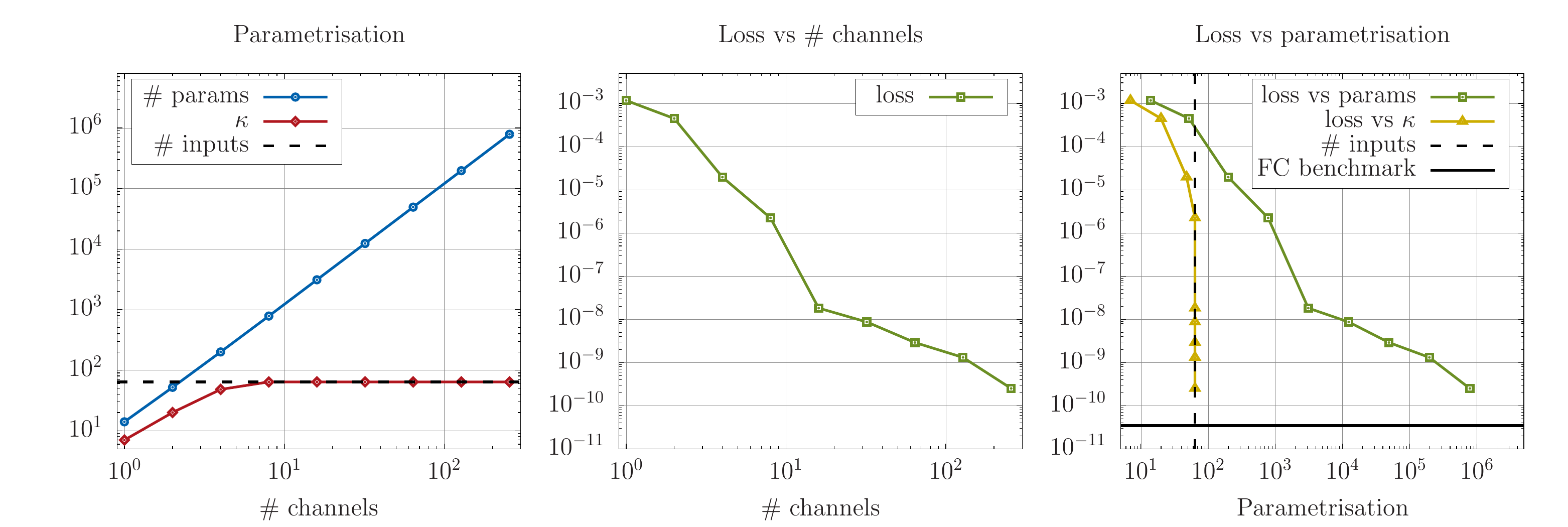}
\centering
\caption{\label{fig:capacity} \textit{(left)} In red, the total capacity of hierarchical models with $L=6, n_l = 2, d_l=2^{l-1}, c_l \in \{1, 2, 4, 8, 16, 32, 64, 128, 256\}$. 
The dashed line corresponds to $\kappa = n$, such that the class of models is equal to the whole space $\mathbb{R}^n$. The blue line represents the total number of parameters, which grows quadratically in the number of channels. \textit{(center)} The optimal loss found by a L-BFGS-B 2nd-order optimizer, which roughly decreases as a power law of the number of channels. \textit{(right)} The same loss as a function of the number of parameters (green) and effective parameters (yellow). The benchmark corresponds to the numerical  loss obtained for a fully connected model (FC). 
}
\end{figure}

Fig. \ref{fig:capacity} plots the total capacity of hierarchical models with parameters $L=6, n_l = 2, d_l=2^{l-1}$ and a variable number of channels $c_l \in \{1, 2, 4, 8, 16, 32, 64, 128, 256\}$ as a function of its total number of parameters.\footnote{The number of effective parameters can be easily computed for $c_l = 1$ and $d_l=n_l^{l-1}$. In that case, it is equal to $ \sum_{l=1}^L (n_l - 1) + 1$, whereas the total number of parameters is $\sum_{l=1}^L n_l$.} Although the total number of parameters scales quadratically with the number of channels, the number of effective parameters scales more slowly until it reaches the upper boundary $\kappa=n$, where the whole space becomes accessible and the exact model $A^*$ can be attained (note that the equality is exact, as $\kappa$ is  by definition an integer).  As the graph on the right shows, the loss decreases as a power law of the number of parameters, then saturates when it reaches the loss obtained for a reference model with one parameter per input (i.e. a fully connected model). Interestingly, the transition happens beyond the point where $\kappa=n$ is first reached, due to numerical errors: the optimization process seems to be more efficient when the model is overparametrized.

\begin{figure}[!t]
\hspace*{-1cm} \includegraphics[width=1.1\textwidth]{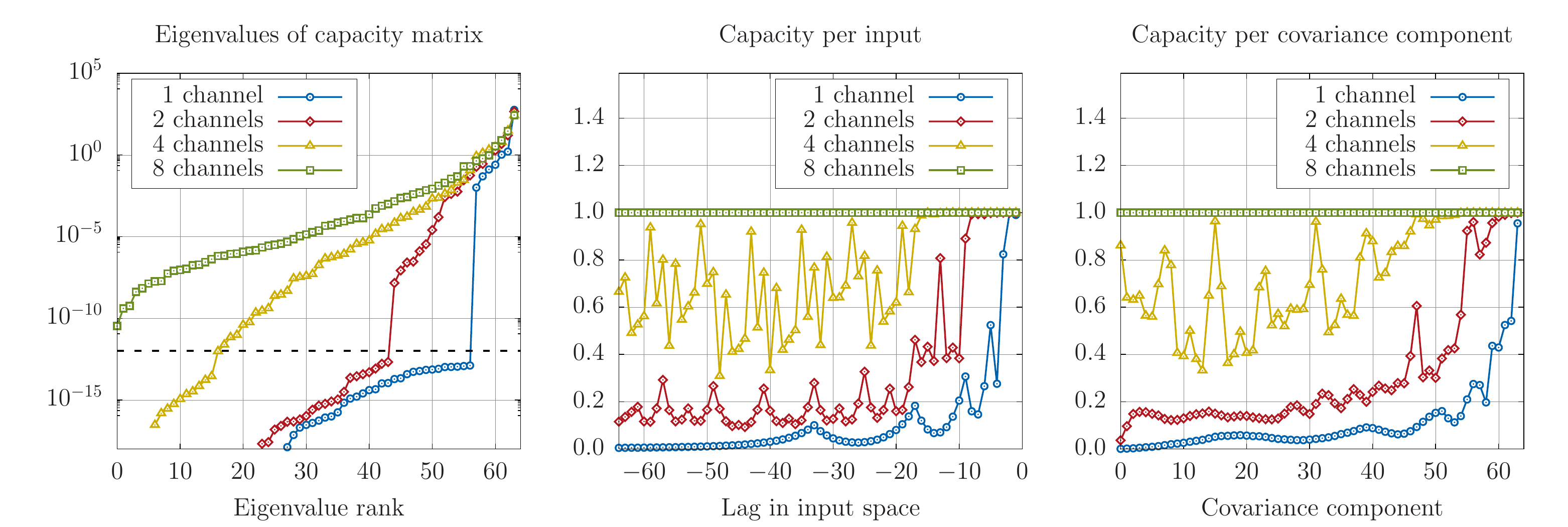}
\centering
\caption{\label{fig:spatial_capacity} \textit{(left)} The eigenvalues of  the capacity weighting matrix $\Lambda$, for a number of channels $c_l \in \{1, 2, 4, 8\}$. The dashed line is the threshold that was used to separate truly zero from non-zero eigenvalues. \textit{(center)} The spatial capacity allocation for the same models. The 8-channels model allocates a capacity of 1 for all dimensions. \textit{(right)} The capacity allocation along the eigenvectors of the covariance matrix of the process.}
\end{figure}

The spatial capacity allocation $\{\kappa_{s_1}, ..., \kappa_{s_n}\}$ along the natural basis of the input space for the same models is shown in Fig. \ref{fig:spatial_capacity}. The left plot represents the eigenvalues of the \emph{capacity weighting matrix} $\Lambda$ defined in Section \ref{sec:constraints}, and whose number of non-zero values corresponds to the model capacity $\kappa$. Because the optimization process has a finite precision, $\Lambda$ typically doesn't have any zero eigenvalues, but in practice it is often possible to separate small but genuinely non-zero eigenvalues from noisy ``zero'' eigenvalues.\footnote{In particular, noise-induced non-zero eigenvalues are typically symmetric around zero, whereas genuinely nonzero eigenvalues are always positive. The distribution of negative eigenvalues can therefore be used to find the scale of the noise on the positive half-space.} The plot in the middle represents the capacity per natural input dimensions, which we also call the \emph{capacity per input} (CPI), and which corresponds to the number of parameters that the model dedicates to modelling direct dependencies on a given input. It is defined as the set $\{\kappa_{s_i} = \|K^Ts_i \|_F^2 \mid i=1,\ldots,n\}$ where $K$ is the orthonormal basis of the space of constraints and $s_i$ is the one-hot vector corresponding to the input at distance $i$. In this example, more capacity is allocated to the recent past (e.g. $\kappa_{s_1}$, on the very right) than on the distant past (e.g. $\kappa_{s_n}$, on the very left). As the total capacity increases with the number of channels, so does the spatial capacity per input dimension. Notably, as the capacity increases, the shortest range dependencies are fully modelled first. Longer range dependencies are only allocated capacity once shorter dependencies are modelled. Finally, the plot on the right shows the capacity allocated along the eigenvectors of the covariance matrix, which often shows a cleaner pattern but doesn't allow for a spatial interpretation.


\subsubsection{Errors vs. capacity}\label{sec:errors_vs_capacity}

\begin{figure}[!t]
\includegraphics[width=0.9\textwidth]{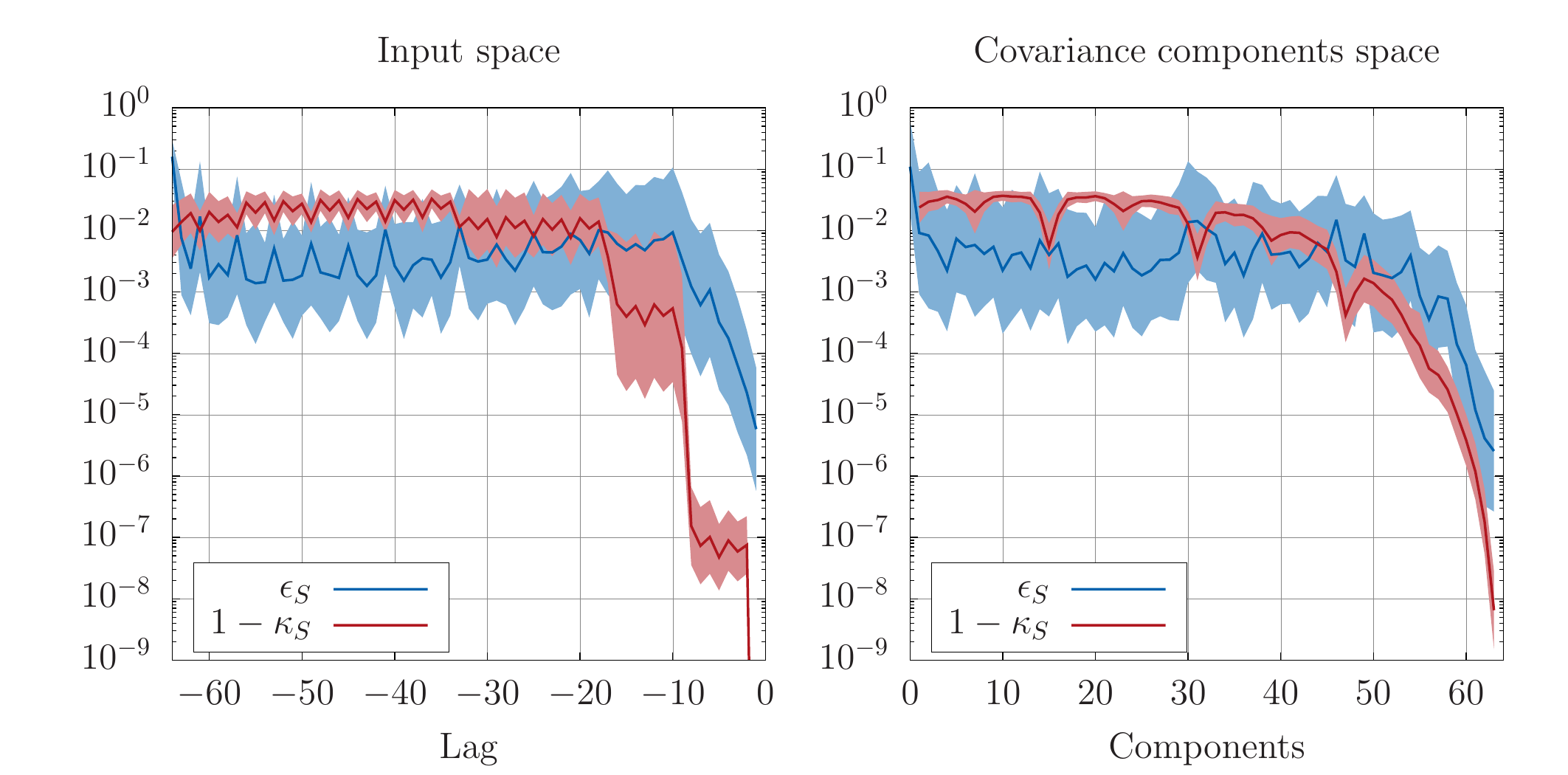}
\centering
\caption{\label{fig:errors_vs_capacity} (left) The normalized capacity bounds $1 - \kappa_S$ and squared errors $\epsilon_S$, as a function of the lag. (right) Idem, along the covariance eigenvectors. The capacity bound has a much smaller variance than the realized errors.}
\end{figure}

We can compare the squared errors along the input space dimensions (i.e. the AR coefficients) with the bound  $n_S - \kappa_S$ from Eq. (\ref{eq:error_bound}), to evaluate the relationship empirically (with $n_S = 1$ along each input dimension). Figure \ref{fig:errors_vs_capacity} shows the average (plain line) as well as the standard deviation (colored area) across many runs of the optimization process for randomly initialized models with $c_l = 4$. Since the relationship between the capacity bound and the squared errors is defined up to some constant, both have been normalized to sum to 1. The figure confirms qualitatively the relationship from Eq. (\ref{eq:error_bound}):  $\epsilon_S = O\left(n_S - \kappa_S \right)$. The relationship appears to hold quite accurately along the covariance components -- better than along the input dimensions. 
In general, the capacity bound is much more stable across runs than the realized squared errors, which makes it a good candidate for analyzing an architecture in a more intrinsic way.

\subsubsection{Further analysis}\label{sec:wavenet_further}

One conclusion from the spatial capacity analysis conducted in Figure \ref{fig:spatial_capacity} is that the hierarchical structure tends to focus the model capacity on short range dependencies for the process considered, at the expense of long range structure. Could we dissect this behaviour layer by layer?

\begin{figure}[!t]
\includegraphics[width=0.9\textwidth]{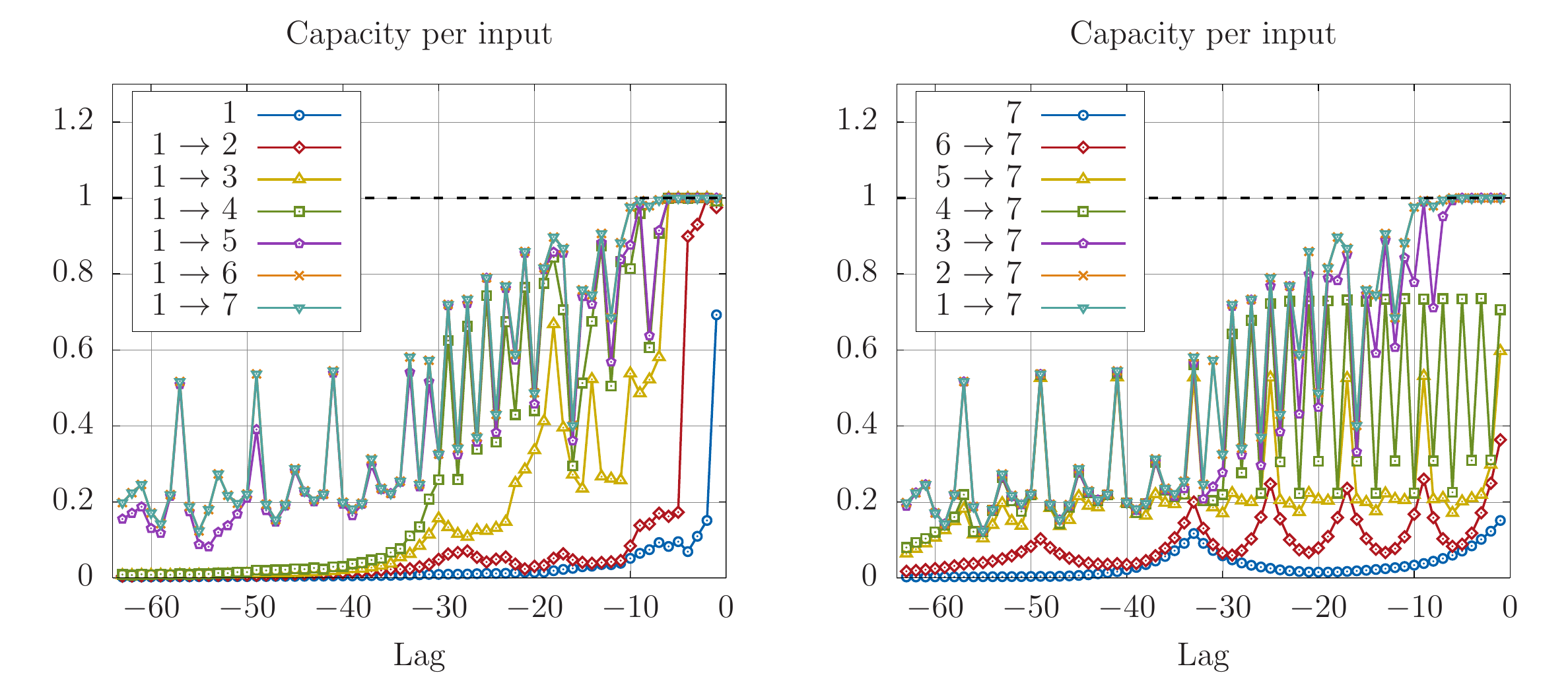}
\centering
\caption{\label{fig:conditional_capacity_chain} Conditional capacity chain in (left) forward order of the layers and (right) backward order.}
\end{figure}

We first use the conditional capacity defined in Section \ref{sec:conditional_capacity} to evaluate the contribution of each layer to the total model capacity,  for a hierarchical model with $L=6, n_l = 2, d_l=2^{l-1}, c_l =4$. Figure \ref{fig:conditional_capacity_chain} shows the chained conditional capacity contributions of the model layers, illustrating Property \ref{prop:conditional}. Such analysis requires to choose an arbitrary order for the layers: here we compare the forward order (i.e. starting from the lowermost layer) with the backward order (starting from the uppermost layer).  The figures show that most of the short-term capacity allocation is realized by the few lowermost layers, while most of the long term capacity allocation is realized by the uppermost layers -- as expected. Put differently, the lowermost layers end up allocating their capacity for reproducing the short term dependencies -- because they can, and that it's optimal for the prediction task. 
In light of these results, it seems unlikely for example that such model trained on audio will learn wavelet-like filters (which would be uniformly useful across the input space). Instead, their modelling capacity will be allocated to extracting signal from the recent past, insofar as possible. In the context of audio modelling, this encourages the use of two-scale models, with one part of the model trying to capture short-term dependencies, and the other part trying to capture more universal features.

\begin{figure}[!t]
\hspace*{-1cm} \includegraphics[width=\textwidth]{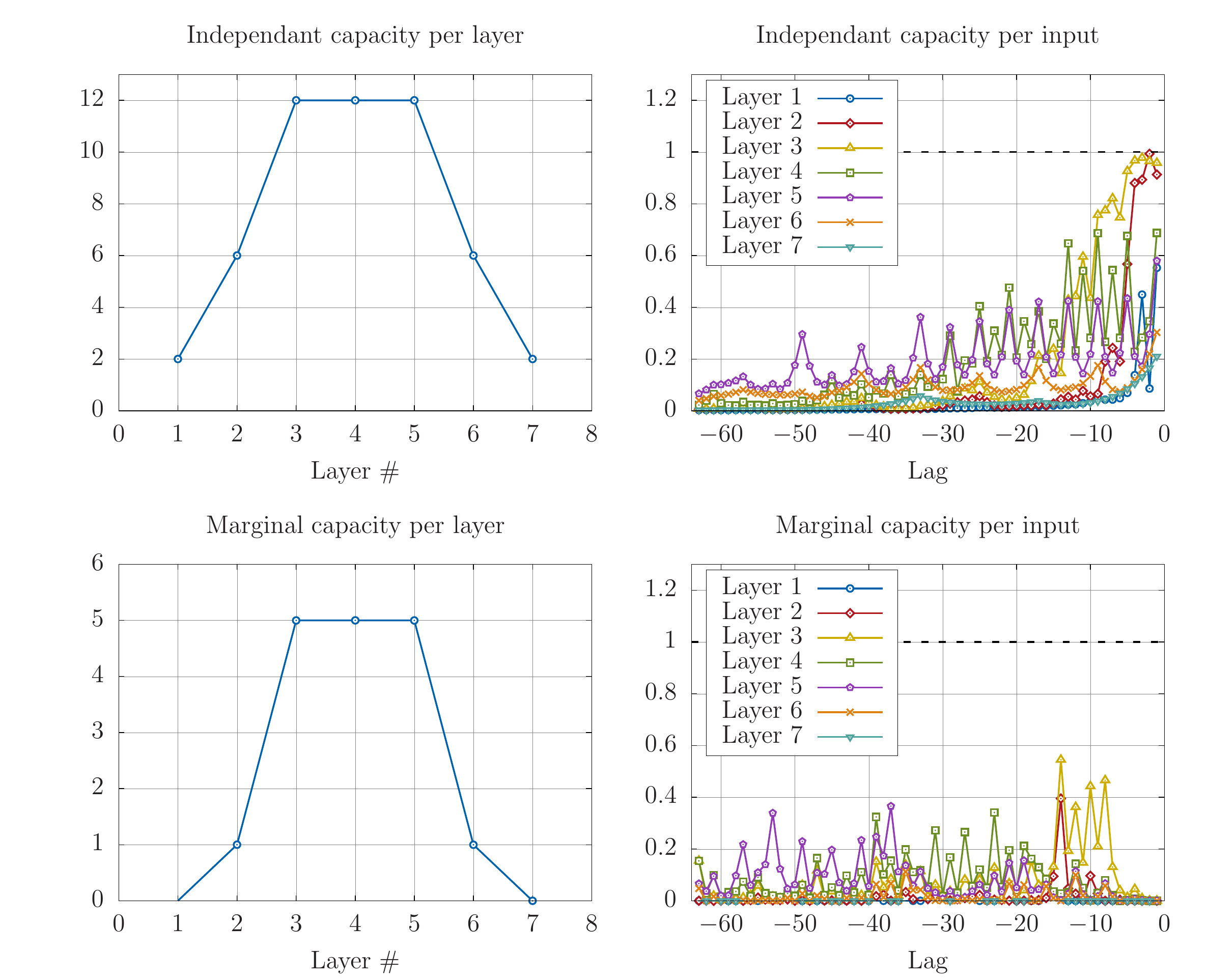}
\centering
\caption{\label{fig:marginal_capacities} (top) Independent capacity allocation of each layer in the hierarchical network. On the left, the total capacity, on the right, the capacity per input. (bottom) The marginal contributions of these layers, conditioned to the rest of the model. On the left, the total capacity, on the right, the capacity per input. }
\end{figure}

Figure \ref{fig:marginal_capacities} finally quantifies how each layer behaves individually, in two ways: (i) by analyzing its capacity allocation independently from all other layers (i.e. as if all other layers' parameters were constants), and (ii) by analysing its \emph{marginal contribution} to the total model's capacity allocation, defined as their conditional capacity given the space of constraints associated with all other parameters in the model. 
The observations are consistent with Figure \ref{fig:conditional_capacity_chain}: the lower the layer, the more the capacity allocation is peaked around the recent past, while higher layers tend to achieve a more uniform allocation. Analyzing the marginal contributions is also interesting. Naturally, the marginal contributions are lower than the independent contributions as the former is some residual of the latter. More specifically, it seems that the middle layers are the one that are the least redundant, while some layers have a zero or near-zero marginal contribution to the model capacity allocation (more on this in Appendix \ref{sec:redundancy}).








\subsection{Recurrent models}

As an alternative to hierarchical models for audio modelling, \cite{mehri2016samplernn, kalchbrenner2018efficient} have used recurrent models as a way to encode dependencies between inputs that are arbitrarily far apart (using some architecture and back-propagation tricks to make training manageable). In this section, we analyze the simplest linear recurrent models, and compare their behaviour to that of the hierarchical models of the previous section.

\subsubsection{Model definition}

\begin{figure}[t]

\def\layersep{1.7}
\def\nodesep{2}
\def\nodesize{12pt}
\def\margins{40}

\definecolor{neworange}{RGB}{255,178,102}
\definecolor{newblue}{RGB}{62, 138, 255}

\newcommand\mydots{\makebox[3em][c]{.\hfil.\hfil.}}

\hspace*{2.8cm}  \begin{tikzpicture}[shorten >=1pt,->,draw=black!80, node distance=\layersep,line width=0.5mm]
    \tikzstyle{every pin edge}=[<-,shorten <=1pt]
    \tikzstyle{neuron}=[circle,draw=black!80,minimum size=\nodesize,inner sep=0pt,line width=0.2mm]
    \tikzstyle{input neuron}=[neuron,fill=yellow!50];
    \tikzstyle{invisible neuron}=[neuron,minimum size=0pt];
    \tikzstyle{output neuron}=[neuron,fill=red!60];
    \tikzstyle{hidden neuron}=[neuron];
    \tikzstyle{annot} = [ text centered]
    
        \node[invisible neuron] (i1) at (-0.2 * \nodesep,0) {};
        \node[neuron, fill=newblue!80] (n1) at (1*\nodesep,0) {};
        \node[neuron, fill=newblue!80] (n2) at (2*\nodesep,0) {};
        \node[neuron, fill=newblue!80] (n3) at (3*\nodesep,0) {};
        \node[neuron, fill=newblue!80] (n4) at (4*\nodesep,0) {};
        
        \node[invisible neuron] (i5) at ((0,\layersep) {};
        \node[neuron] (n5) at (1*\nodesep,\layersep) {};
        \node[neuron] (n6) at (2*\nodesep,\layersep) {};
        \node[neuron] (n7) at (3*\nodesep,\layersep) {};
        \node[neuron] (n8) at (4*\nodesep,\layersep) {};
        
        \node[invisible neuron] (i9) at ((-0.2 * \nodesep,2*\layersep) {};
        \node[neuron, fill=neworange!80] (n9) at (1*\nodesep,2*\layersep) {};
        \node[neuron, fill=neworange!80] (n10) at (2*\nodesep,2*\layersep) {};
        \node[neuron, fill=neworange!80] (n11) at (3*\nodesep,2*\layersep) {};
        \node[neuron, fill=neworange!80] (n12) at (4*\nodesep,2*\layersep) {};

          \path[-] (i1) -- node[auto=false]{\mydots} (n1);
          \path[line width = 0.4mm,-{Latex[length=2.7mm,width=2.2mm]}]  (i5) edge (n5);
          \path[-] (i9) -- node[auto=false]{\mydots} (n9);
         

   \path[line width=0.4mm, -{Latex[length=2.7mm,width=2.2mm]}] (n1) edge (n5);
   \path[line width=0.4mm, -{Latex[length=2.7mm,width=2.2mm]}]  (n2) edge (n6);
   \path[line width=0.4mm, -{Latex[length=2.7mm,width=2.2mm]}]  (n3) edge (n7);
   \path[line width=0.4mm, -{Latex[length=2.7mm,width=2.2mm]}]  (n4) edge (n8);
   
   \path[line width=0.4mm, -{Latex[length=2.7mm,width=2.2mm]}]  (n5) edge (n6);
   \path[line width=0.4mm, -{Latex[length=2.7mm,width=2.2mm]}]  (n6) edge (n7);
   \path[line width=0.4mm, -{Latex[length=2.7mm,width=2.2mm]}]  (n7) edge (n8);

   \path[line width=0.4mm, -{Latex[length=2.7mm,width=2.2mm]}]  (n5) edge (n9);
   \path[line width=0.4mm, -{Latex[length=2.7mm,width=2.2mm]}]  (n6) edge (n10);
   \path[line width=0.4mm, -{Latex[length=2.7mm,width=2.2mm]}]  (n7) edge (n11);
   \path[line width=0.4mm, -{Latex[length=2.7mm,width=2.2mm]}]  (n8) edge (n12);

    \node[annot,right of=n4, anchor=west, align=justify, text width=4cm, node distance=\margins] (ny1) {\small\textbf{Input}};
    \node[annot,right of=n8, anchor=west, align=justify,  text width=4cm, node distance=\margins] (ny2) {\small\textbf{Hidden layer}};
    \node[annot,right of=n12, anchor=west, align=justify, text width=4cm, node distance=\margins] (ny3) {\small\textbf{Ouptut}}; 
    
       \node[annot] (Win) at (1.3*\nodesep, 0.5 * \layersep)  {$W^{(in)}$} ;
       \node[annot] (Win) at (2.5*\nodesep, 0.5 * \layersep)  {\dots} ;
       \node[annot] (Win) at (4.3*\nodesep, 0.5 * \layersep)  {$W^{(in)}$} ;
       
       \node[annot] (Wout) at (1.3*\nodesep, 1.5 * \layersep)  {$W^{(out)}$} ;
       \node[annot] (Wout) at (2.5*\nodesep, 1.5 * \layersep)  {\dots} ;
       \node[annot] (Wout) at (4.3*\nodesep, 1.5 * \layersep)  {$W^{(out)}$} ;
       
       \node[annot] (Wout) at (0.5*\nodesep, 1.2 * \layersep)  {$W^{(rec)}$} ;
       \node[annot] (Wout) at (3.5*\nodesep, 1.2 * \layersep)  {$W^{(rec)}$} ;

\end{tikzpicture}

\caption{\label{fig:recurrent_schema} The architecture of a recurrent model with one hidden layer.}

\end{figure}
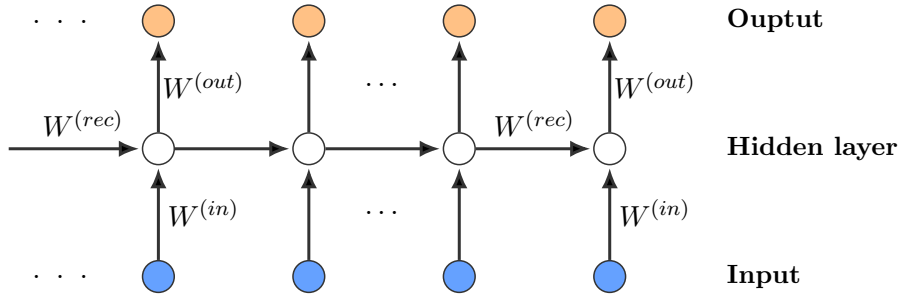

For the purpose of this study, we consider one particular type of linear recurrent models with a single recurrent layer (cf. Figure \ref{fig:recurrent_schema}), and whose number of parameters scales linearly with the number of channels $c$. More precisely, the space of models $\mathcal{A}$ is parametrized as:

\begin{equation}\label{eq:space_recurrent}
\mathcal{A} = \left\{ \left(a_i = \sum_{j=1}^{c} W^{(\mathrm{in})}_j \left(W^{(\mathrm{rec})}_j\right)^{n-i} W^{(\mathrm{out})}_j \right) _{i=1..n} \mid  W^{(\mathrm{in})}, W^{(\mathrm{rec})}, W^{(\mathrm{out})} \in \mathbb{R}^{c} \right\},
\end{equation}

\noindent where $W^{(\mathrm{in})}$ and $W^{(\mathrm{out})}$ are 1x1 convolutions and $W^{(\mathrm{rec})}$ is a recurrent layer with no links across channels.

\subsubsection{Capacity analysis}

\begin{figure}[!t]
\hspace*{-1cm}\includegraphics[width=1.08\textwidth]{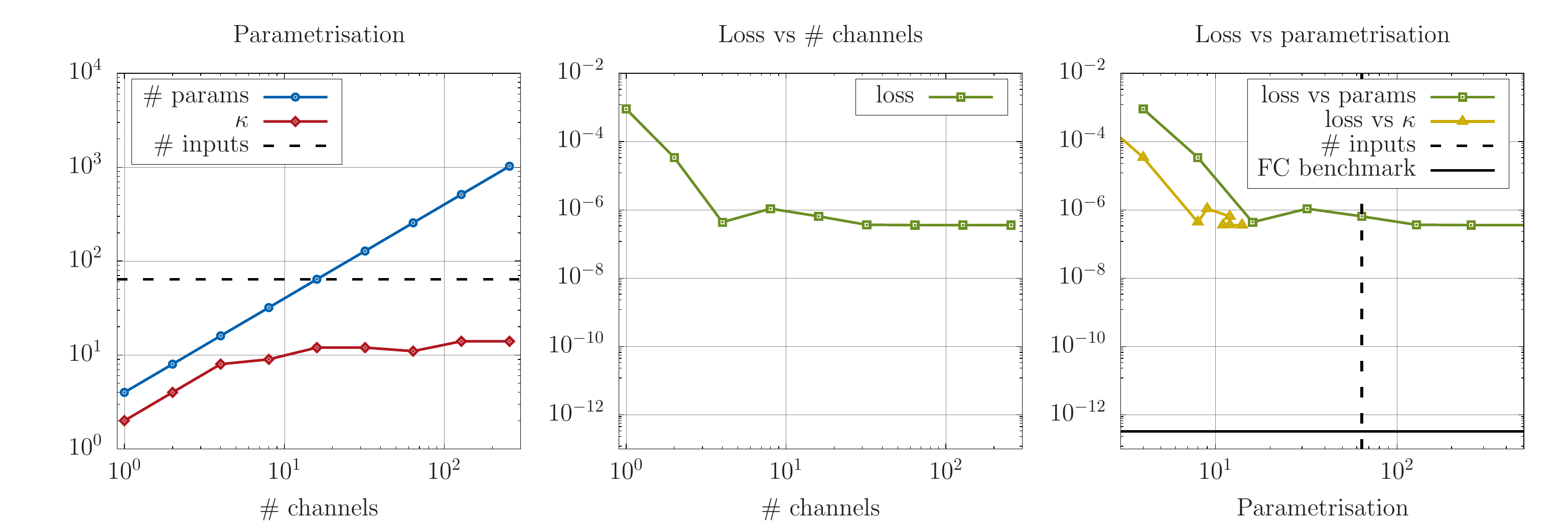}
\centering
\caption{\label{fig:recurrent_capacity} \textit{(left)} In red, the total capacity of a recurrent models with $c \in \{1, 2, 4, 8, 16, 32, 64, 128, 256\}$. 
The blue line represents the total number of parameters, which grows linearly in the number of channels. The recurrent model never reaches full capacity in this case (dashed line). \textit{(center)} The optimal loss found by a L-BFGS-B 2nd-order optimizer. \textit{(right)} The same loss as a function of the number of parameters (green) and effective parameters (yellow). The benchmark corresponds to the numerical  loss obtained for a fully connected model (FC) 
}
\end{figure}

\begin{figure}[!t]
\hspace*{-1cm}\includegraphics[width=1.08\textwidth]{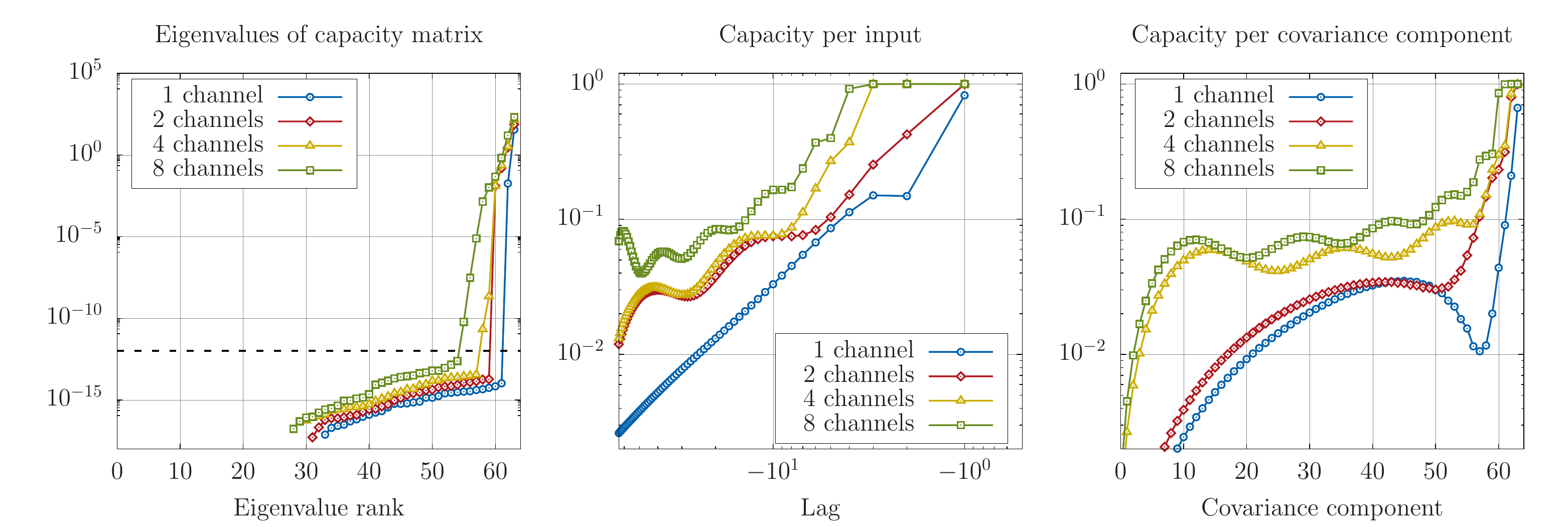}
\centering
\caption{\label{fig:recurrent_spatial_capacity} \textit{(left)} The eigenvalues of  the capacity weighting matrix $\Lambda$, for a number of channels $c \in \{1, 2, 4, 8\}$. 
\textit{(center)} The corresponding spatial capacity allocation, which decreases as a power law of the spatial distance. \textit{(right)} The capacity allocation along the eigenvectors of the covariance matrix of the process.}
\end{figure}

As in the previous section, we analyze the model capacity as a function of its number of parameters. We vary the number of channels $c \in \{1, 2, 4, 8, 16, 32, 64, 128, 256\}$, and plot the corresponding number of parameters and effective parameters (i.e. the total capacity) in Figure \ref{fig:recurrent_capacity}. We also plot the loss as a function of the number of channels and as a function of the number of (effective) parameters. In this case, it appears that the loss decreases as a power law of the number of effective parameters. Finally, Figure \ref{fig:recurrent_spatial_capacity} shows the corresponding spatial capacity analysis, for $c \in \{1, 2, 4, 8\}$. As above, the model shares its capacity between (i) a few close inputs, to which it allocates a capacity of 1, and (ii) more distant inputs, to which it allocates a power-law decreasing capacity.

\section{Towards richer models}\label{sec:multidim}

\subsection{Multi-dimensional inputs}

The measure of a model's capacity allocation introduced above only makes sense in linear contexts - linear models, linear processes. To make a first step towards richer models, we now consider multi-dimensional gaussian input processes (of dimension $d$). In a way, this is just a remake of everything that was presented in the previous sections -- but with richer dependencies between the inputs and the variable to predict. 

The spatial capacity analysis is of particular interest, as the subspace $\mathcal{S}$ corresponding to inputs at a given spatial position is now $d$-dimensional. The maximum capacity allocation for one spatial position will thus also be equal to $d\geq 1$. 
Therefore, for $d$ large enough, it should be rarer to reach the degenerate situations where the capacity saturates at its maximal value for short-term dependencies (as seen in Figure \ref{fig:spatial_capacity}). Rather, by increasing the dimensionality of the inputs, one should expect to observe different tradeoffs between short-term and long-term capacity allocation.

\begin{figure}[!t]
\centering
\includegraphics[width=0.9\textwidth]{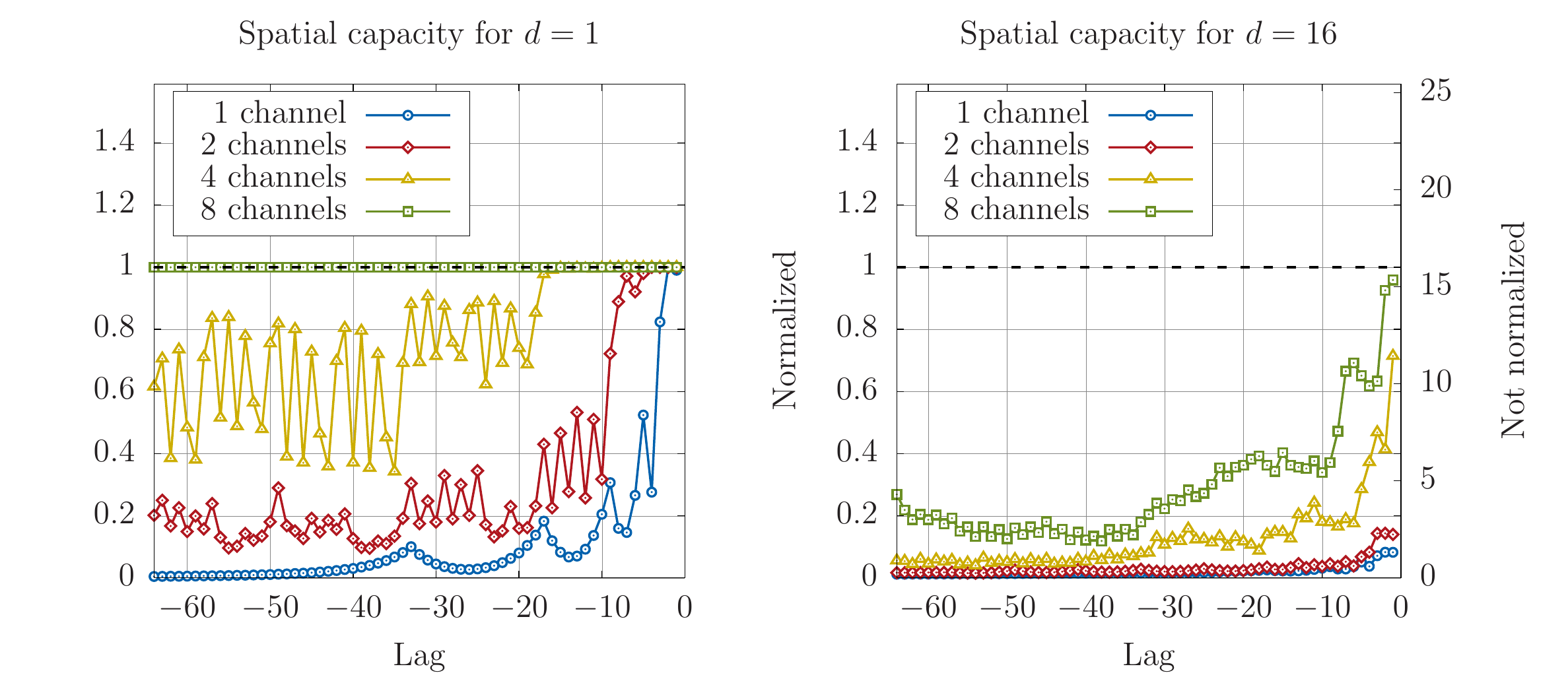}
\caption{\label{fig:multidim16_spatial_capacity_scaling} \textit{(left)} Spatial capacity for $d=1$ and $c_l\in\{1, 2, 4, 8 \}$, similar to Figure \ref{fig:spatial_capacity}.\textit{(right)}  Spatial capacity for $d=16$ and $c_l\in\{1, 2, 4, 8 \}$. The left axis corresponds the capacity normalized by $d$, such that reaching the value $1$ for a given lag means that the dependencies at that distance are perfectly modelled. The right axis corresponds to the capacity itself (i.e. quantifying the number of parameters used for each lag).}
\end{figure}

The first analysis, which we present in Figure \ref{fig:multidim16_spatial_capacity_scaling}, compares the scaling of the capacity with the number of channels, for $d \in \{1, 16\}$ where the $d$ input components are taken to be independent processes with similar autocorrelations. For the 1-dimensional process, we observe the same pattern as in Section \ref{sec:examples_hierarchical}, where a full capacity is first allocated to the most short-term dependencies, then spills over to longer ones as the ceiling $\kappa_S = 1$ is reached. For the 16-dimensional process, more capacity continues to be allocated to short-term dependencies beyond $\kappa_S = 1$. For a similar number of parameters, more capacity is therefore allocated for modelling the recent past when the relationships between input and output are more complex.

\begin{figure}[!t]
\centering
\hspace*{-1cm}\includegraphics[width=1.08\textwidth]{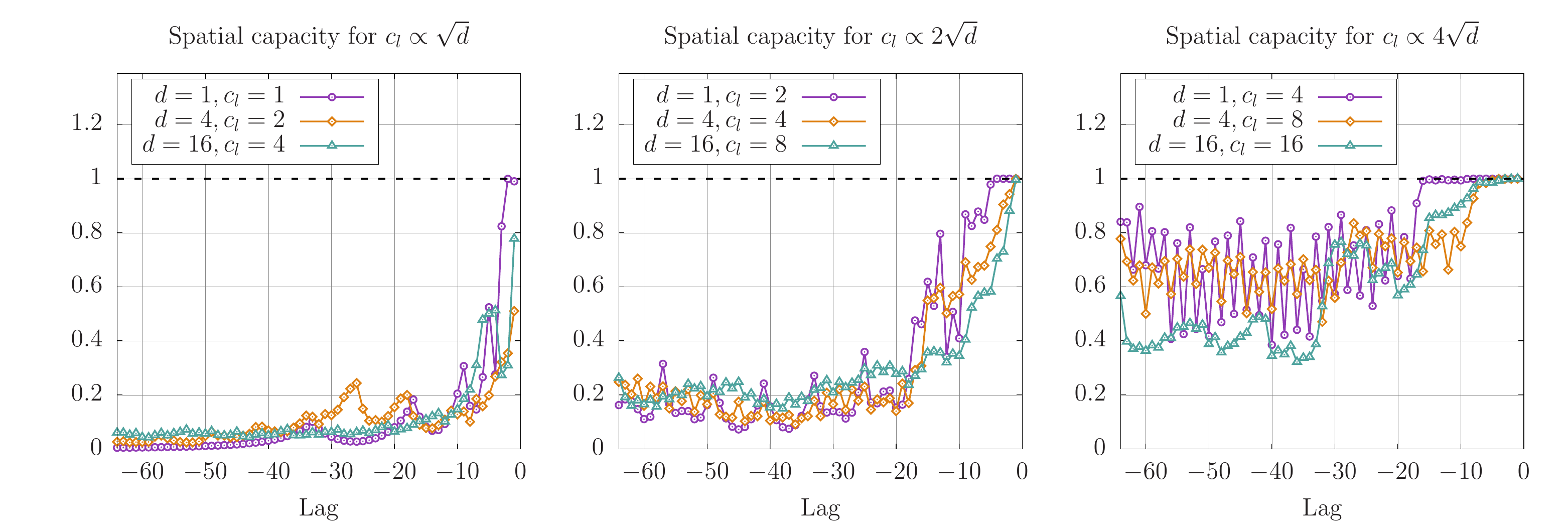}
\caption{\label{fig:non_linearity_channels_scaling}  Spatial capacity for $d\in\{1, 4, 16 \}$ and\textit{(left)} $c_l = \sqrt{d}$,  \textit{(center)} $c_l = 2\sqrt{d}$ and  \textit{(right)} $c_l = 4\sqrt{d}$. For a given architecture, the normalized spatial capacity allocations mostly depends on the ratio between the number of parameters and the input dimensionality (i.e. $c_l^2 / d$ here).}
\end{figure}

Can we quantify better the interplay between $d$ and the number of parameters in the model ? Figure \ref{fig:non_linearity_channels_scaling} shows the capacity allocation normalized by the dimensionality of the input process, when the number of parameters is scaled proportionally to the dimensionality of the input (equivalently, the number of channels $c_l$ is scaled as $\sqrt{d}$). As one might have expected, the relative capacities are almost equivalent -- only smoother in the higher dimensional case. This suggests that one can study the capacity allocated for high dimensional processes and large number of parameters, simply by scaling down the dimensionality of the input and the number of parameters proportionally. 

\subsection{Non-linear models}

\subsubsection{Feature space}

A simple instance of non-linear problems are those where the prediction is a linear function of some fixed non-linear feature map applied to the input $Y$. The space of such functions is defined as:

\begin{equation}\label{eq:feature_map}
\begin{aligned}
\varphi_W: \quad& \mathbb{R}^n & \to &  \mathbb{R}\\
 & Y &  \mapsto & A_W^T\phi(Y),
\end{aligned}
\end{equation}

\noindent where $\phi: \mathbb{R}^n \to \mathbb{R}^m$ is some fixed function that maps the input $Y\in\mathbb{R}^n$ to some feature space $\mathbb{R}^m$ and $A_W\in \mathbb{R}^m$ is a linear model that acts on the feature space. As in Section \ref{sec:applying_capacity}, $A_W$ is an element of the space of linear functions $\mathcal{A}$  defined by some mapping from some parameter space:

\begin{equation}
\begin{aligned}
 f: \quad& \mathbb{R}^p & \to & \mathbb{R}^m\\
    & W & \mapsto & A_W
\end{aligned}
\end{equation}

\noindent The $L^2$ loss function is then:

\begin{equation}
\begin{aligned}
\mathcal{L} &= \mathbb{E}\left[ \left(A - A^*\right) ^T \phi(Y) \phi(Y)^T \left(A - A^*\right)  \right]\\
&= \left(A - A^*\right) ^T   \Sigma^\phi  \left(A - A^*\right),
\end{aligned}
\end{equation}

\noindent where $\Sigma^\phi :=  \mathbb{E}\left[ \phi(Y) \phi(Y)^T \right]$. In the trivial case $\phi(Y) = Y$, one recovers exactly the setting of Section  \ref{sec:applying_capacity}. In general, $\phi$ can be non-linear and $m$ can be arbitrarily large, leading to a much richer set of functions than considered above. 

\subsubsection{Capacity allocation in the feature space}

Because of the linear nature of the problem in the feature space, one can apply capacity analysis as previously \emph{in the feature space}, by substituting 

\begin{equation}
\begin{aligned}
&\Sigma \to \Sigma^\phi\\
&\mathbb{R}^n \to \mathbb{R}^m.
\end{aligned}
\end{equation}

\noindent As above, one can compute $\tilde{K}^\phi :=  \Sigma^\phi \frac{\partial A}{\partial W} \in \mathbb{R}^{m\times p}$, find an orthonormal basis $K^\phi$ of $\mathrm{Vect}(\tilde{K}^\phi)$ and define the capacity allocated to a subspace $\mathcal{S}^\phi$ \emph{of the feature space}:

\begin{equation}
\kappa^\phi(\mathcal{S}^\phi) = \| \left(K^\phi\right)^T S^\phi  \|_F^2.
\end{equation}

\subsubsection{Capacity allocation in the input space}

The remaining challenge is then to define a notion of capacity \emph{in the input space}. While the task does not appear to be straightforward in general, there is one special case where the question is simpler: when $\phi$ acts on the different input components separately, such that we can write:

\begin{equation}
\phi(Y) = \left(\phi^1(y_1), ... ,\phi^d(y_1), ..., \phi^1(y_n), ...\phi^d(y_n)\right)
\end{equation}

\noindent where $\left\{ \phi^j\right\}_{1\leq j \leq d}$ are linearly independent functions (for example, polynomial basis functions, Fourier basis, etc.). If we denote by $\left(\kappa^1_1, ... ,\kappa^d_1, ..., \kappa^1_n, ...\kappa^d_n\right)$ the capacities corresponding to the natural dimensions of the feature space, then the capacity allocated to the $i$-th input component can be written as:

\begin{equation}
\kappa_i = \sum_{j=1}^d \kappa^j_i.
\end{equation}

\noindent Just like in the multi-dimensional case of the previous section, the maximum capacity per input component is $d$, as one parameter per basis function $\phi^j$ is now necessary to fully model the dependencies. The size $d$ of the set of basis functions defines the complexity of the data dependencies -- which is typically infinite for real data. This illustrates the fact that the notion of under-parametrization becomes much more common as the data complexity increases, and so does the regime in which capacity analysis makes sense.

The above analysis is only a glimpse of how the notion of capacity generalizes in non-linear settings. A more thorough study in the context of non-linear neural network layers is presented in \cite{donier2018nonlinear}.




\section{Conclusion}

In this paper, we have introduced the notion of capacity analysis for linear systems. We have defined a linear model's capacity $\kappa$, which represents the number of independent parameters that describe the model space, and shown how this capacity can be broken down along input subspaces. In particular, we have focussed on spatial capacity allocation along natural dimensions of the input space. We have illustrated these concepts in the case of 1-dimensional hierarchical and recurrent models, and shown that some typical allocation patterns arise for each type of architecture. Finally, we have made a step towards capacity allocation in richer settings, by considering multi-dimensional inputs and non-linear feature maps. This opens the door for more principled network design, by going beyond the value of the loss function and better understanding which dependencies a given architecture can be expected to capture. This is only a first step towards a deeper theoretical understanding of neural networks through the lens of capacity allocation, and the journey ahead is still long. One obvious next step is to perform capacity analysis across a number of architectural variants, and see if or how this can guide us through architecture design. But to be really useful, the concept of capacity analysis first needs to be generalized to other non-linear models -- for example, non-linear neural networks. 

\section*{Acknowledgements} The author would like to thank Martin Gould, Marc Sarfati and Antoine Tilloy for their very useful comments on the manuscript.


\bibliography{Architecture_arxiv_submission}{}
\bibliographystyle{plain}

\clearpage
\appendix



\section{Proof of Property \ref{prop:sum}}\label{sec:proof}

First note that $\forall K_1, K_2, ~\| \left[ K_1 \mid K_2 \right]^T S  \|_F^2 = \| K_1^T S  \|_F^2  + \| K_2^T S  \|_F^2$ and therefore the left hand side is equivalent to $(\forall \mathcal{S},~\| K^T S  \|_F^2 =\| \left[ K_1 \mid K_2 \right]^T S  \|_F^2)$.
Since $\mathcal{K}=\mathcal{K}_1+ \mathcal{K}_2$ and $K, K_1, K_2$ have orthonormal columns, one can write  $\left[ K_1 \mid K_2 \right] = K R$ where $R\in\mathbb{R}^{\kappa\times\left(\kappa_1+\kappa_2 \right)}$ has unitary columns such that $\mathrm{Tr} \left( RR^T \right) = \kappa_1+\kappa_2$. Note that $R$ is a square rotation matrix if and only if $\mathcal{K}_1 \obot \mathcal{K}_2$, therefore one needs to prove the following equivalence:

\begin{equation}
(\forall \mathcal{S},~\| R^T K^T S  \|_F^2 =\| K^T S  \|_F^2) \quad \Leftrightarrow \quad R \text{ rotation matrix}.
\end{equation}

\noindent If $R$ is a rotation matrix, then the left hand side proposition is trivially true by invariance property of the Frobenius norm. Let us now prove the forward implication and assume that the left hand side proposition holds. Since the columns of $K$ are orthonormal vectors, then $\forall X\in\mathbb{R}^\kappa, \exists S\in\mathbb{R}^n \mid K^T S = X$. Therefore, $\forall X\in\mathbb{R}^\kappa,~\| R^T X  \|_F^2 =\| X \|_F^2$ which implies that $RR^T = \mathbb{I}_{\mathbb{R}^\kappa}$. Taking the trace, one obtains $\kappa_1+\kappa_2 = \kappa$, which shows that $\mathcal{K}=\mathcal{K}_1 \bigoplus \mathcal{K}_2$. $R$ is therefore a square matrix that verifies $RR^T = \mathbb{I}_{\mathbb{R}^\kappa}$, i.e. a rotation matrix.

\section{Gaussian processes and linear models}\label{sec:GP}

\subsection{Gaussian process prediction}\label{sec:prediction}

Gaussian processes (GP) are processes $\{y_i\}$ for which the joint distribution of any finite set of points is gaussian, and which can thus be fully characterized by their mean and covariance matrix. If the process is stationary, we can assume that the mean is zero without loss of generality, and the covariance matrix takes a simple symmetric shape where all diagonals are constant:

\begin{equation}
p(y_1, ..., y_n) = \mathcal{N}\left( 0, \Sigma_n \right), \quad (\Sigma_n)_{ij} = \mathbb{E}(y_i y_j) =: c_{\mid i-j \mid},
\end{equation}

\noindent where the function $\tau \to c_\tau$ is called the autocovariance function of the gaussian process. In this framework, the conditional distribution of a sample $x_{n+1}$ conditioned on some samples $Y_{1:n} := \{y_1, ..., y_{n} \}$ takes the simple form:

\begin{equation}
\begin{aligned}\label{eq:AR_coeffs}
p(y_{n+1} | y_1, ..., y_{n}) =  \mathcal{N}\left(A^* Y_{1:n}, v^* \right), \quad  \quad & A^* := \Sigma_{1:n +1}\Sigma_{n}^{-1}, \\
& v^*:= 1 - \Sigma_{1:n +1}^T\Sigma_{n}^{-1}\Sigma_{1:n +1}
\end{aligned}
\end{equation}

\noindent where $\Sigma_{n}$ is the autocorrelation matrix defined above and we have used the notation $\Sigma_{1:n +1} := ( c_1, ... , c_{n + 1} )$. The best prediction of $y_{n + 1}$ given $\{y_1, ..., y_{n}\}$ is therefore realized by a linear model with coefficients $A^*$, and the corresponding residual variance is $v^*$. Conversely, one can compute the autocovariance function of a gaussian process generated by the linear auto-regressive model $y_{n+1} = A Y_{1:n} + \eta $ where $\eta \sim  \mathcal{N}(0, v)$, by reversing Eq. \ref{eq:AR_coeffs}. Naturally, if one uses $A = A^*$ and $v=v^*$, then one recovers the autocovariance matrix $\Sigma_n$.

\subsection{The optimization problem}\label{sec:optimization}

Let us consider a gaussian process of autocovariance matrix $\Sigma_n$ and the class of linear models:

\begin{equation}
f_A: Y_{1:n} \to \tilde{y}_{n + 1} := A Y_{1:n}, \quad A \in \mathcal{A} \subset \mathbb{R}^n.
\end{equation}

\noindent We want to find the optimal parameter $\hat{A}$ that solves the following optimization problem:

\begin{equation}
\begin{aligned}\label{eq:problem}
\hat{A} &= \underset{A\in \mathcal{A}}{\text{argmin}}~ \text{Var}\left[ \tilde{y}_{n + 1} -  y_{n + 1} \right]\\
 &= \underset{A\in \mathcal{A}}{\text{argmin}} \left(A - A^*\right) ^T \Sigma_n \left(A - A^*\right) + v^* =: \mathcal{L}
\end{aligned}
\end{equation}

\noindent where $A^*$ is the optimal prediction from Section \ref{sec:optimization}, and the solution to this problem for $\mathcal{A} = \mathbb{R}^n$. In general, for underparametrized models, $A^*$ has no reason \textit{a priori} to be in $\mathcal{A}$, in which case $\hat{A} \neq A^*$ and the residual variance is $\hat{v} > v^*$.

The above optimization problem has no general solution as the model space $\mathcal{A}$ can be arbitrarily complex, as we have seen in Section \ref{sec:examples}. However, thanks to the linearity of the problem, the residual variance to be minimized could be expressed more directly using the autocorrelation matrix $\Sigma_n$, allowing to eliminate of the stochasticity of the problem and perform a more stable and straightforward gradient descent (instead of a stochastic gradient descent). This enables us to use a second order optimization which finds a near-optimal solution in seconds, even for models that have millions of parameters.

\subsection{The hierarchical example}\label{sec:hierarchical_example}

\begin{figure}[!ht]
\centering
\hspace*{-1cm}  \includegraphics[width=1.08\textwidth]{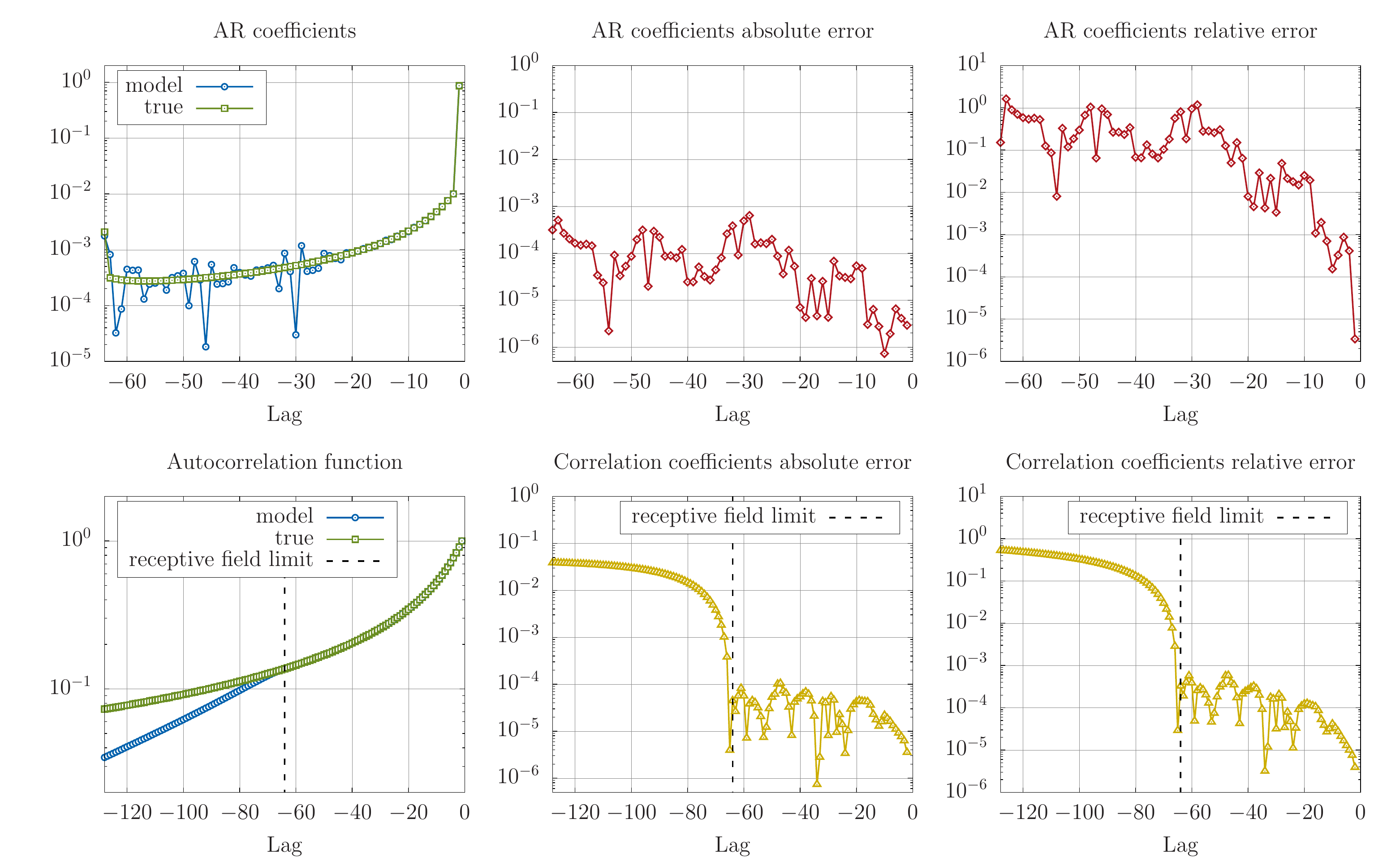}
\caption{\label{fig:coeffs} \textit{(top left)} The model AR parameters (in blue) vs. the true parameters (in green).  \textit{(top center and right)} The absolute and relative difference between model coefficients and true coefficients. \textit{(bottom left)} The model autocorrelation (in blue) vs. the true autocorrelation (in green).  \textit{(bottom center and right)} The absolute and relative differences between model autocorrelation and and true autocorrelation.
}
\end{figure}

One can compare the optimal parameters $\hat{A}$ with the true parameters $A^*$ and the corresponding autocorrelation $\hat{C}$ with the true autocorrelation $C$. The curves as well as their relative and absolute differences are shown in Fig. \ref{fig:coeffs} for a number of channels $c_l = 4$.

We can make the two following observations from the plots:

\begin{enumerate}[(i)]

\item The error on the autocorrelation $\|\hat{C} - C \|$ is lower for short lags. This behaviour is expected as $\hat{C} - C $ is obtained by solving the same linear system as for obtaining $\hat{C}$ from $\hat{A}$, but with a boundary condition $(\hat{C} - C)_0 = 0$ instead of $\hat{C}_0 = 1$.

\item The error on the coefficients $\|\hat{A} - A \|$ shows a less clear pattern. There is in fact a competition between two forces: the model tends to allocate more capacity for reproducing the largest target coefficients  (as observed in Fig. \ref{fig:spatial_capacity}), but they are also more difficult to reproduce exactly. Depending on the dominant force, larger coefficients will be either better approximated (in absolute terms) or not. The capacity theory developed in the previous sections is in fact a good way to isolate and measure the first force. Indeed, as we've seen in Fig. \ref{fig:spatial_capacity}, the patterns in the capacity plots are much cleaner and much more interpretable than the realized errors of Fig. \ref{fig:coeffs}.



\end{enumerate}





\section{Wavenets and repeated layers}\label{sec:wavenet_repeat}

\begin{figure}[!ht]
\centering
\includegraphics[width=0.6\textwidth]{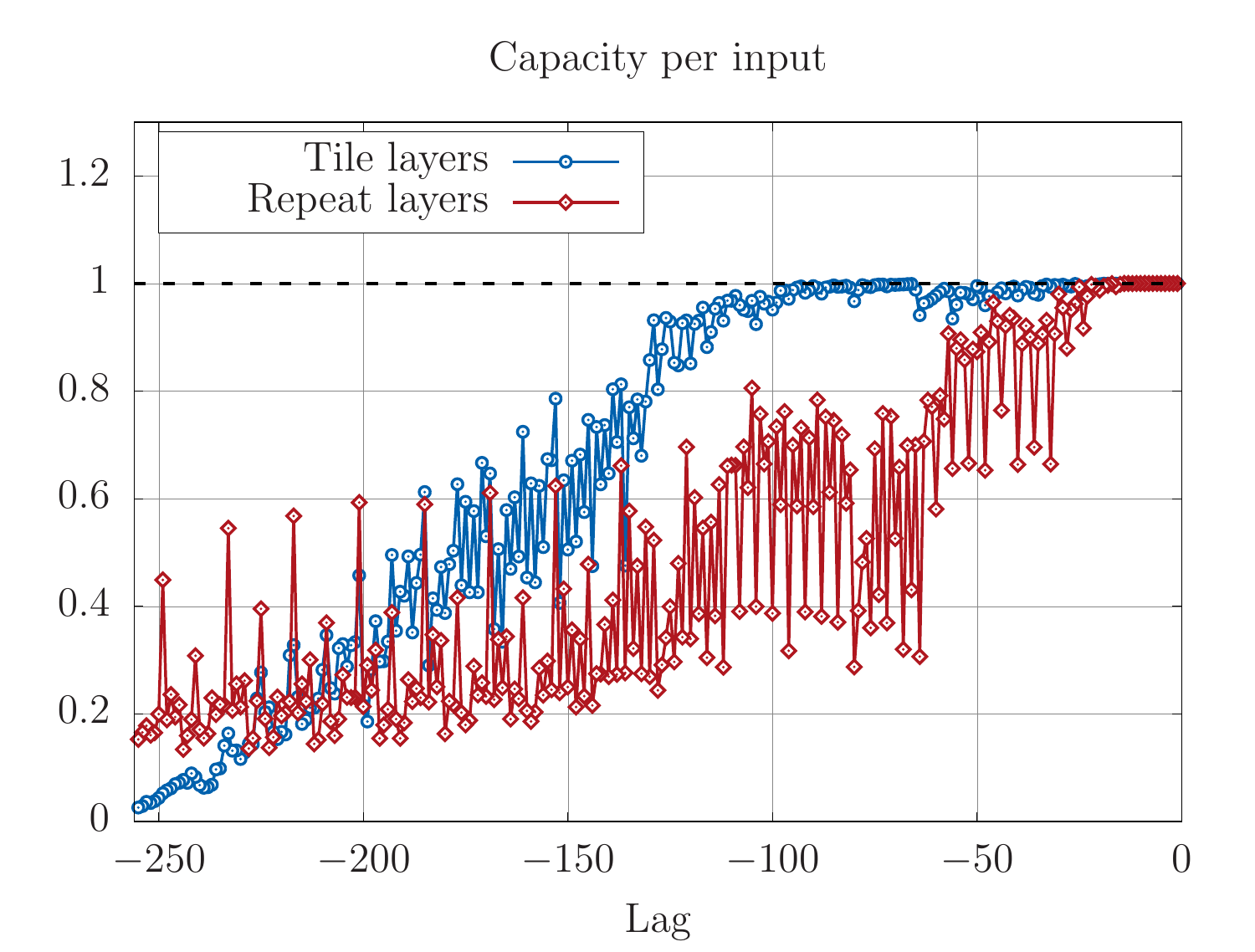}
\caption{\label{fig:tile_vs_repeat}  Spatial capacity allocation for (blue) tiled blocks with dilation pattern $\{1, 2, ..., 128, 1, 2, ..., 128, 1, 2, ..., 128\}$ and (red) repeated layers with dilation pattern $\{1, 1, 1, 2, 2, 2, ..., 128, 128, 128\}$.}
\end{figure}

One architecture trick that was introduced in \cite{van2016wavenet} is to \emph{tile} dilated blocks, with a dilation pattern that looks like $\{1, 2, 4, ..., 128, 1, 2, 4, ..., 128, 1, 2, 4, ..., 128\}$. One interesting question is how such architectural choice differs from \emph{repeating} layers instead of blocks: $\{1, 1, 1, 2, 2, 2, 4, 4, 4, ..., 128, 128, 128\}$. Figure \ref{fig:tile_vs_repeat} show the spatial capacity allocations for the two variants aforementioned and a number of channels $c_l=4$, for the same task and the same data. Perhaps as expected, the capacity allocation patterns are highly different: the tiled version has a larger total capacity for the same number of parameters, and allocated most of it to short range dependencies. On the contrary, the repeated version has a lower total capacity (more redundant parameters), but puts more focus on the distant past.

\section{Layer redundancy}\label{sec:redundancy}

In Section \ref{sec:conditional_capacity}, we have defined the concept of conditional capacity, and we have considered some examples in the case of hierarchical models in Section \ref{sec:wavenet_further}. We have observed in particular that some layers have a marginal capacity equal to 0. Here we formulate some hypotheses regarding such layers with zero marginal capacity.

\begin{definition}
Let $W\in \mathbb{R}^p$ be a set optimal parameters wrt some optimization criterion, $\tilde{W}$ be a subset of these parameters and $W_{\setminus \tilde{W}}$ be the set of all parameters $W$ except $\tilde{W}$. Let $\mathcal{K}$ and $\mathcal{K}_{\setminus \tilde{W}}$ denote the space of constraints respectively associated to $W$ and $W_{\setminus \tilde{W}}$. Then, $\mathcal{K} = \mathcal{K}_{\setminus \tilde{W}}$ if and only if the marginal contribution of $\tilde{W}$ to the capacity is zero for every subspace. In this case, the parameters $\tilde{W}$ are said to be \emph{(jointly) redundant}.
\end{definition}

The above property defines what it means for a set of parameters to be jointly redundant: namely, the constraints associated with such parameters could be ignored without affecting the capacity allocation. When this is the case, we are making two conjectures regarding these parameters:

\begin{conjecture}
If a set of parameters is redundant, then for almost all values of these parameters, the optimal model can be recovered by adapting the other parameters.  The parameter values for which this does not hold are those that lead to degenerate cases, i.e. to spaces of constraints of lower dimensions. The measure of such set is zero.
\end{conjecture}

\begin{conjecture}
If a set of parameters is redundant, then with probability 1 these parameters can be fixed at random before learning the other parameters, without affecting the optimal model.
\end{conjecture}

The first conjecture comes from the intuition that if a parameter is redundant, any change in this parameter can be compensated by tweaking other parameters. For example, if the model space is $\mathcal{A} = \{w_1 w_2 A_0 \mid (w_1, w_2)\in\mathbb{R}^2\}$, and if one optimal model is found with parameters $w_1^*$ and $w_2^*$, then for any fixed $w_2\neq 0$, the optimal model can be recovered for $w_1 = w_1^* w_2^* / w_2$. The case $w_2= 0$ is the only degenerate case in this example. The second conjecture immediately follows from the first, as the set of degenerate parameters has measure 0. Although redundant parameters are more likely to happen in linear settings, it has been observed that fixing a large fraction of the weights at random in deep networks might result in performance that is on par with fully learnable models \cite{rosenfeld2018intriguing, saxe2011random}, which might be related to the above conjectures.





\end{document}